\documentclass[preprint,12pt]{elsarticle}

\usepackage{amssymb}
\usepackage{amsmath}
\usepackage{xcolor}
\usepackage{booktabs}
\usepackage{colortbl}
\usepackage{natbib}
\usepackage{bm}
\usepackage{subfigure}
\usepackage{rotating}
\usepackage{comment}
\usepackage{lineno}
\usepackage{hyperref}
\usepackage{multirow}

\journal{xx}

\begin{document}

\begin{frontmatter}



\title{LT-PINN: Lagrangian Topology-conscious Physics-informed Neural Network for Boundary-focused Engineering Optimization} %% Article title


\author[PolyU-CEE]{Yuanye Zhou} %% Author name
\author[PolyU-ME,PolyU-SFT]{Zhaokun Wang} %% Author name
\author[PolyU-CEE,PolyU-RISUD]{Kai Zhou\fnref{1}}  %% Author name 
\author[PolyU-ME]{Hui Tang} %% Author name
\author[HKU-ME]{Xiaofan Li}

\address[PolyU-CEE]{Department of Civil and Environmental Engineering, The Hong Kong Polytechnic University, Hong Kong, China}
\address[PolyU-RISUD]{Research Institute for Sustainable Urban Development (RISUD), The Hong Kong Polytechnic University, Hong Kong, China}
\address[PolyU-ME]{Department of Mechanical Engineering, The Hong Kong Polytechnic University, Hong Kong, China}
\address[PolyU-SFT]{School of Fashion and Textiles, The Hong Kong Polytechnic University, Hong Kong, China}
\address[HKU-ME]{Department of Mechanical Engineering, The University of Hong Kong, Hong Kong, China}

\fntext[1]{Corresponding author: \texttt{cee-kai.zhou@polyu.edu.hk}}

% \linenumbers

%% Abstract
\begin{abstract}

Physics-informed neural networks (PINNs) have emerged as a powerful meshless tool for topology optimization, capable of simultaneously determining optimal topologies and physical solutions. However, conventional PINNs rely on density-based topology descriptions, which necessitate manual interpolation and limit their applicability to complex geometries. To address this, we propose Lagrangian topology-conscious PINNs (LT-PINNs), a novel framework for boundary-focused engineering optimization. By parameterizing the control variables of topology boundary curves as learnable parameters, LT-PINNs eliminate the need for manual interpolation and enable precise boundary determination. We further introduce specialized boundary condition loss function and topology loss function to ensure sharp and accurate boundary representations, even for intricate topologies. The accuracy and robustness of LT-PINNs are validated via two types of partial differential equations (PDEs), including elastic equation with Dirichlet boundary conditions and Laplace’s equation with Neumann boundary conditions. Furthermore, we demonstrate effectiveness of LT-PINNs on more complex time-dependent and time-independent flow problems without relying on measurement data, and showcase their engineering application potential in flow velocity rearrangement, transforming a uniform upstream velocity into a sine-shaped downstream profile. The results demonstrate (1) LT-PINNs achieve substantial reductions in relative $L_2$ errors compared with the state-of-art density topology-oriented PINNs (DT-PINNs), (2) LT-PINNs can handle arbitrary boundary conditions, making them suitable for a wide range of PDEs, and (3) LT-PINNs can infer clear topology boundaries without manual interpolation, especially for complex topologies.

\end{abstract}

\begin{highlights}
\item The proposed LT-PINN is a meshless Lagrangian topology-conscious PINN for topology optimization;
\item It eliminates the error-prone manual boundary interpolation required in density topology-orient PINN approach;
\item It precisely encodes arbitrary boundary conditions through boundary condition loss function, leading to a more accurate prediction of PDE solution;
\item It includes topology loss function to efficiently handle complex topology system with self-similar topology patterns;
\item It generates manufacturable complex topologies directly compatible with CAD.
\end{highlights}

%% Keywords

\end{frontmatter}

%\linenumbers


\section{Introduction}
\label{sec:intro}
%% Labels are used to cross-reference an item using \ref command.
Topology optimization is a design optimization technique that determines the ideal material layout, within a specified design space, to meet desired performance criteria, while satisfying constraints related to boundary conditions \cite{bendsoe1989optimal}. It has been widely applied to many industry applications \cite{pedersen2006industrial}, such as aerospace \cite{zhu2016topology}, automotive \cite{yang1995automotive}, energy \cite{huang2007topology}, and bioengineering \cite{wu2021advances}. While, topology optimization is a computationally intensive process, as it usually requires hundreds of design iterations. In each iteration, the physical field must be solved to obtain the sensitivity information needed to update the current design. Large-scale topology optimization problems are extremely computationally intensive, as they may involve millions or more design variables \cite{mukherjee2021accelerating}. To manage the substantial computational demands, several classical techniques are developed in the past decades, such as parallel computing \cite{borrvall2001large}, \cite{paris2013parallel}, advanced iterative solvers \cite{amir2010efficient}, multi-scale or multi-resolution methods \cite{kim2000multi}, \cite{park2015multi}.

Recently, machine learning has emerged as a rapidly developing paradigm for topology optimization, offering faster and more efficient solutions compared to classical techniques \cite{shin2023topology}, \cite{kim2021machine}, \cite{zhang2024enhancing}, \cite{senhora2022machine}, \cite{yin2024dynamically}. However, purely data-driven machine learning models lack the ability to encode physical information, which can lead to overfitting. In contrast, physics-informed neural networks (PINNs) are especially attractive for topology optimization because they integrate physical information directly into the model. This integration allows for simultaneous determination of optimal topology and solution of physical fields, eliminating the need for iterative processes required by classical techniques and mitigating the overfitting problem in data-driven models. Another notable advantage of PINNs in topology optimization is that they are meshless methods, which are capable of capturing fine geometric details with significant changes in shapes and topologies during the optimization process. So far, PINNs have been successfully applied to solve a wide range of partial differential equations (PDEs) problems \cite{karniadakis2021physics}, such as heat problems \cite{cai2021physics}, high-speed compressible flow \cite{mao2020physics}, vortex-induced vibration \cite{cheng2021deep} and linear elastic problems \cite{wang2024m}.

In a prior study of topology optimization, Chen et al. \cite{chen2020physics} utilized PINN to address the inverse design of photonic metamaterials. They innovatively set both the permittivity map and the PDE solution as outputs of the PINN, enabling the network to simultaneously reconstruct the permittivity distribution and predict the PDE solution. This approach effectively integrated the physical constraints of the unknown metamaterial topology, i.e. permittivity map, into the neural network framework. In a subsequent study, Lu et al. \cite{lu2021physics} proposed a hard-constrained PINN which exactly imposes Dirichlet and periodic boundary conditions into the PINN to improve model accuracy and convergence speed. For structural topology optimization, Jeong et al. \cite{jeong2023complete} developed a density topology-orient PINN that integrates density field into neural network and was able to handle topology optimization without measurement data. However, the aforementioned PINNs for topology optimization predominantly depend on density-based descriptions, such as permittivity maps, density fields, and solid fractions. These descriptions, being inherently continuous outputs of PINNs, only provide an implicit representation of the actual topology. As a result, manual interpolation is required to identify the topology boundary, which complicates human understanding and hinders the engineering reproduction process. Additionally, current PINNs are restricted to Dirichlet boundary conditions because they are unable to accurately determine the boundary location until the optimization process is finalized.

To address the limitation on implicit topology representations, several techniques have been proposed to enhance the capabilities of PINNs based on sharpened density-based descriptions. Jeong et al. \cite{jeong2023complete} applied a high-order power scaling factor on density field to boost its convergence towards either 0 or 1, so as to identify topology boundary. Zhu et al. \cite{zhu2025physics} used a sigmoid activation function on the density field to minimize the buffer region between fluid and solid topology. Mowlavi \& Kamrin \cite{mowlavi2023topology} suggested using Eikonal regularization to enforce a uniform density field everywhere in order to ensure consistency of physical laws across the topology boundary. Yin et al. \cite{yin2024dynamically} utilized Gaussian interpolation to obtain a more accurate material property from the density field. While these techniques indeed enhanced the contrast of boundaries, they were unable to alter the inherently continuous nature of density-based descriptions. This limitation can impede the accuracy of topology generation and necessitate additional manual parameter tuning.

In contrast, to provide a clear topology boundary, some classical topology optimization techniques employ a different approach, known as the Lagrangian topology description \cite{zhang2016lagrangian}. The Lagrangian topology description utilizes a set of moving morphable patches to construct the topology. It is based on the concept that a series of overlapping, simple, and parameterizable patches can be used to represent an object with highly complex topology. This approach provides an explicit method for topology optimization and has led to significant advancements in structural design \cite{li2024comprehensive}, \cite{lei2019machine}, \cite{du2022efficient}, \cite{lotfalian2025integrating}, \cite{raze2021explicit}. For instance, Guo et al. \cite{guo2017self} used B-spline curve to describe the topology boundary of the void and optimize void distribution and shapes. However, classical Lagrangian topology optimization usually requires special meshing schemes, such as adaptive mesh scheme \cite{nguyen2020efficient}, during topology updates, as conventional mesh regeneration for new topologies is computationally expensive.

As meshless methods, PINNs show strong potential for integration with Lagrangian topology descriptions. This synergy could significantly enhance topology optimization efficiency while preserving well-defined boundaries. However, despite this promise, no prior work has successfully merged PINNs with the Lagrangian approach, leaving both the validity and computational advantages of such a hybrid model unexplored. Moreover, achieving seamless integration between these two frameworks presents substantial theoretical and implementation challenges. To bridge this gap and advance PINN-based topology optimization, we propose Lagrangian Topology-conscious PINNs (LT-PINNs), a novel framework that incorporates the control parameters of topology boundary curves as learnable variables. Our approach introduces (1) boundary condition loss function for both Dirichlet and Neumann conditions, enabling precise boundary identification and PDE solution; and (2) topology loss function to handle geometrically complex structures. This algorithmic innovation delivers a new class of PINNs specifically designed for high-fidelity topology optimization, particularly in scenarios requiring explicit boundary representation.

The remaining of the paper is organized as follows: The methodological principles underlying our proposed approach and its benchmark counterpart are thoroughly examined with theoretical details in Sec. \ref{sec1}. A series of numerical experiments spanning diverse physical problem domains are presented and compared in Sec. \ref{sec2} to highlight the strength of the proposed approach, followed by the conclusion given by Sec. \ref{sec:conclusion}.

\section{Methodologies}
\label{sec1}
This section begins with a review of underlying principles for PINNs and density topology-oriented PINNs (DT-PINNs) as the current state-of-the-art in topology optimization. The proposed novel Lagrangian topology-conscious PINNs (LT-PINNs) with key advancement features then is introduced.

\subsection{DT-PINNs}
\label{subsec:DPINN}
For any general partial differential equation (PDE) expressed as: 
\begin{equation}\label{eq:pde}
\mathcal{N}_{\omega}[\bm{u}(\bm{x}, t)] = f(\bm{x},t),~ \bm{x} \in R^D, 
\end{equation}
where $\mathcal{N}$ denotes any differential operator, $\bm{x}$ and $t$ are the spatial and temporal coordinates, respectively, $D$ is the dimension of space, $\bm{u}$ is the solution, $\omega$ is a parameter that defines the differential operator, and $f$ is a known function. 

In forward PDE problems, the parameter $\omega$ is known, and the goal is to solve Eq. \eqref{eq:pde} subject to the given initial and boundary conditions. Conversely, in inverse problems, $\omega$ is unknown. The objective is thus to determine the solution to Eq. \eqref{eq:pde} and to infer the value of $\omega$ based on available data for $\bm{u}$ and the equation. Specifically, for topology optimization, the unknown $\omega$ refers to the topology representation $\rho$. Ideally, it is a Dirac delta function as below:
\begin{equation}\label{eq:rho}
\begin{aligned}
\rho(\bm{x},t) = 
\begin{cases}
0, & \bm{x} \in \Omega_0\\
1, & \bm{x} \in \Omega_1,
\end{cases}
\end{aligned}
\end{equation}  
where $\Omega_1$ denotes the topology occupied domain, and $\Omega_0$ denotes the rest of the domain. Eq. \eqref{eq:pde} is then modified as:
\begin{equation}\label{eq:rhopde}
(1-\rho(\bm{x},t))(\mathcal{N}_{\omega^*}[\bm{u}(\bm{x}, t)] - f(\bm{x},t))=0, 
\end{equation}
where $\omega^*$ represents the known components of $\omega$, such as fluid viscosity, Young's modulus, Poisson's ratio, and so on. 

\begin{figure}[tp]
\centering%% For centre alignment of image.
\subfigure[]{\label{fig:PINN-a}
\includegraphics[width=1\textwidth]{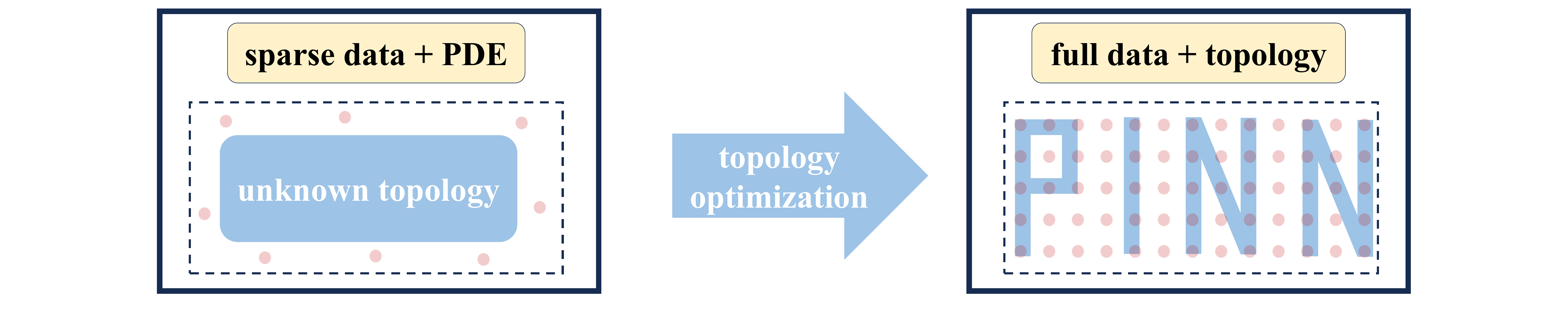}}
\subfigure[]{\label{fig:PINN-b}
\includegraphics[width=1\textwidth]{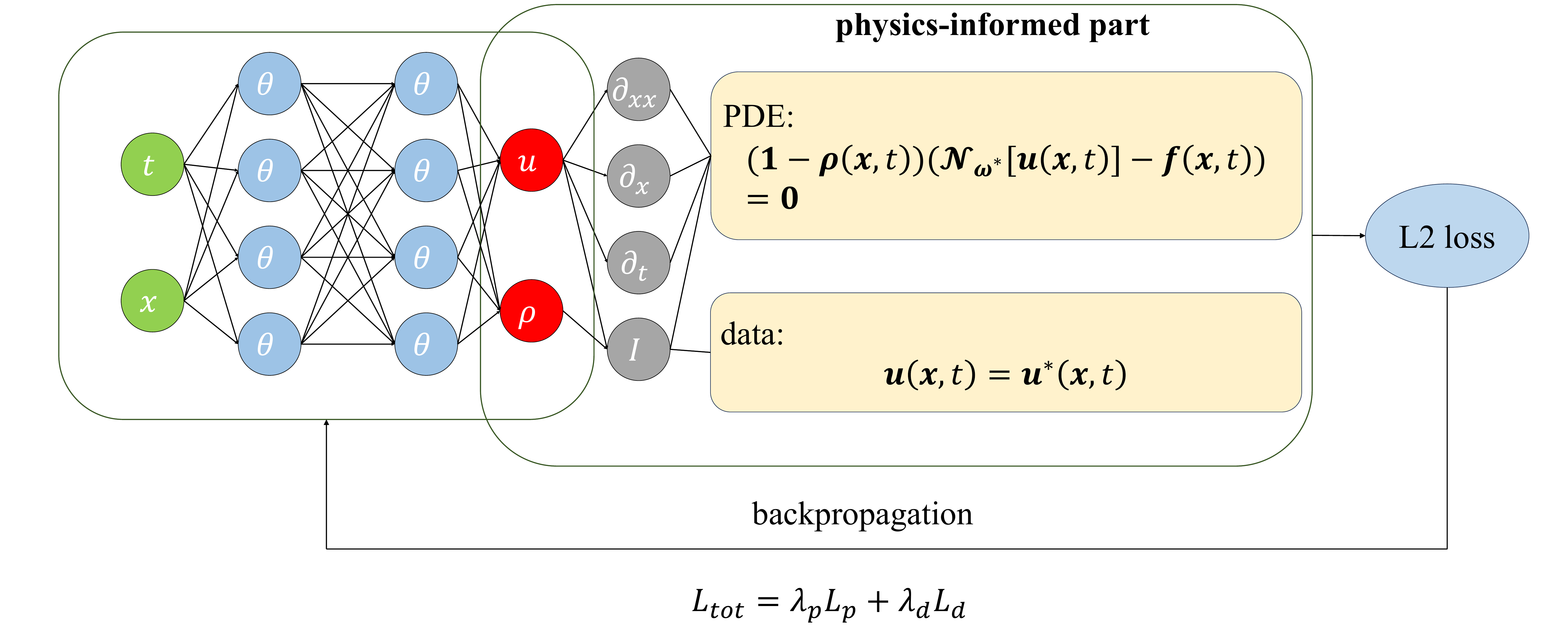}}
\caption{Illustration of topology optimization via DT-PINNs: a) the objective of topology optimization; b) the framework of DT-PINNs.}\label{fig:PINN}
\end{figure}

The objective of topology optimization is then to determine the full solution $\bm{u}$ to Eq. \eqref{eq:rhopde} and infer $\rho$ subject to sparse measurement data of  $\bm{u}$ and PDE, as shown in Fig. \ref{fig:PINN-a}. Before optimization, within the region of interest (ROI), the topology is unknown and only sparse measurement data is known. After topology optimization, both the topology and full solution data of PDE can be determined.

In order to achieve the objective, DT-PINNs are developed as illustrated in Fig. \ref{fig:PINN-b}. In DT-PINNs, there exists a feed-forward neural network (FNN) which takes $\bm{x}$ and $t$ as inputs to approximate the solution to $\bm{u}$ and $\rho$. With the automatic differentiation, the PDE can then be encoded to the neural networks. The total loss function for training the DT-PINNs is expressed as follows:
\begin{equation}
\begin{aligned}
L_{tot}(\bm{\theta}) &= \lambda_{p} L_{p} + \lambda_{d} L_{d},  \\
L_{p} &= \frac{1}{N}\sum^N_{i=1}R_i^2,~ R_i = (1-\rho(\bm{x},t))_i(\mathcal{N}_{\omega^*}[\bm{u}(\bm{x}, t)] - f(\bm{x}, t))_i,\\
L_{d} &=\frac{1}{N_{d}}\sum^{N_{d}}_{i=1}(\bm{u} - \bm{u}^*)^2_i, 
\end{aligned}
\label{losseq1}
\end{equation}
where $L_{tot}$ is the total loss, $L_{p}$ is the PDE loss, $L_{d}$ is the data loss, $\bm{\theta}$ represents the parameters in the neural networks, $R$ is the residual of PDEs, $\bm{u}^*$ is the sparse measurements/initial conditions on $\bm{u}$, $N$ is the number of sample points for PDE loss, and $N_{d}$ is the number of sample points for data loss. $\lambda_{p}$ and $\lambda_{d}$ are the weights of the PDE loss and data loss respectively.  

DT-PINNs encodes physics-information of PDEs via PDE loss ($L_{p}$) in its total loss function ($L_{tot}$). When $L_{tot}$ is zero, it implies that both $\bm{u}$ and $\rho$ are well solved, and topology (represented by $\rho=1$, as defined in Eq. \eqref{eq:rho}) can be inferred. In general, the total loss is minimized during training through stochastic gradient descent optimization. 

Note that in DT-PINNs, the inferred $\rho$ must remain a continuous scalar function to approximate the Dirac delta function, as required by automatic differentiation. Following training completion, a threshold value ($\rho^*$) between 0 and 1 needs to be manually selected to identify the topology boundary. Due to the sharp transition in $\rho$ across this boundary, the resulting topology exhibits significant sensitivity to the chosen threshold value. Furthermore, DT-PINNs struggle to enforce boundary conditions since the topology boundary only becomes discernible after training. While this limitation can be partially mitigated through composite PDE loss function incorporating Dirichlet boundary conditions, as proposed in Zhu et al. \cite{zhu2025physics}, the fundamental challenge remains. The Dirichlet boundary conditions incorporated PDE loss function is mathematically described as: 
\begin{equation}\label{eq:dirichlet}
\begin{aligned}
L_{p} &= \frac{1}{N}\sum^N_{i=1}R_i^2, \\
R_i = & (1-\hat\rho(\bm{x},t))_i(\mathcal{N}_{\omega^*}[\bm{u}(\bm{x}, t)] - f(\bm{x},t))_i ~ + \\
&\hat\rho(\bm{x},t)_i(\bm{u}(\bm{x}, t)-\bm{u_{b}^*}(\bm{x}, t))_i,\\
\hat\rho&=\frac{1}{1+e^{-c\rho}},
\end{aligned}
\end{equation}
where $\hat\rho$ is normalized $\rho$, $c$ is a constant parameter, and $\bm{u_b^*}$ is the known Dirichlet boundary conditions. In this study, we use Eq. \eqref{eq:dirichlet} as the PDE loss function for DT-PINNs with $c=-10$, following the suggestion from Zhu et al. \cite{zhu2025physics}. Nevertheless, such composite PDE loss function cannot accommodate boundary conditions requiring known boundary normal directions, such as Neumann boundary conditions.

\subsection{LT-PINNs}
\label{subsec:LPINN}
\begin{figure}[tp]
\centering%% For centre alignment of image.
\subfigure[]{\label{fig:LTPINN-a}
\includegraphics[width=1\textwidth]{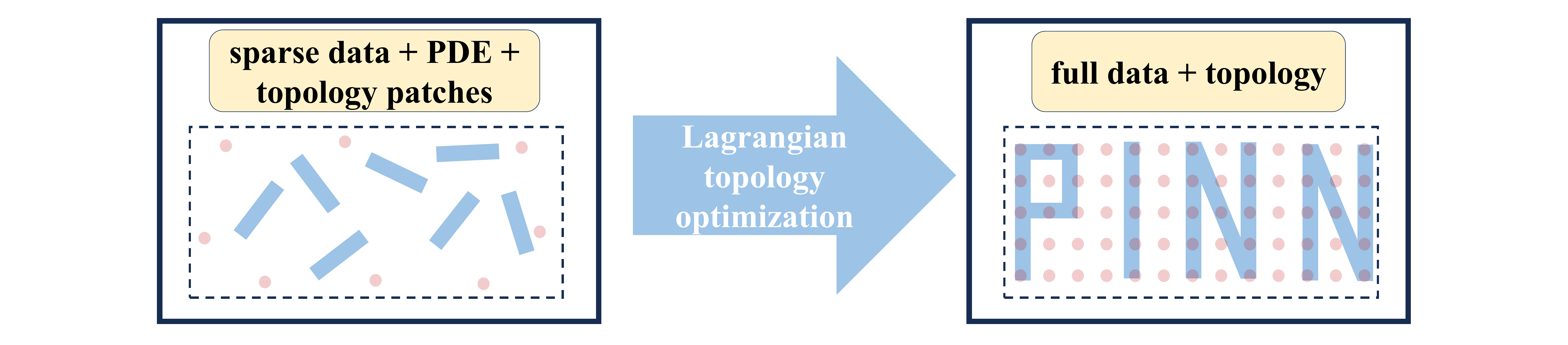}}
\subfigure[]{\label{fig:LTPINN-b}
\includegraphics[width=1\textwidth]{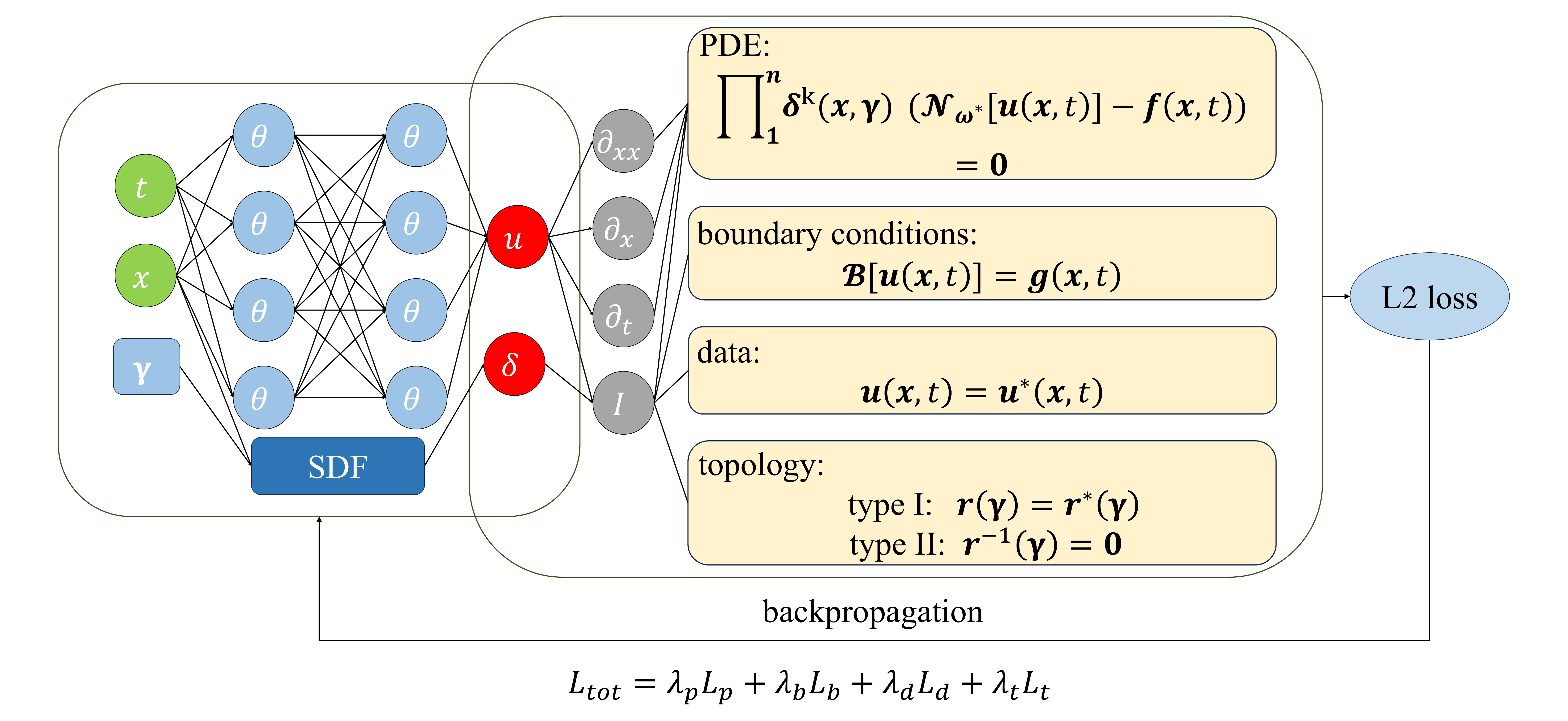}}
\caption{Illustration of topology optimization via LT-PINNs: a) the objective of Lagrangian topology optimization; b) the framework of LT-PINNs.}\label{fig:LTPINN}
\end{figure}

To overcome the key limitations of DT-PINNs, specifically their reliance on manual boundary determination and limited adaptability to boundary conditions, we proposed LT-PINNs built upon Lagrangian topology optimization approach. Its framework is illustrated in Fig. \ref{fig:LTPINN}. This new approach allows one to determine the full solution data of PDEs and infer topology subject to sparse measurement data, PDEs, and prior random distributed topology patches. The fundamental idea relies on constructing complex topologies through compositions of simple, overlapping, and parameterized patches. Since each patch admits an explicit mathematical representation, the topology boundary can be determined in each training step, allowing boundary conditions to be rigorously enforced through corresponding loss functions. 

In LT-PINNs, a geometric distance function (i.e., $\delta(\bm{x},\bm{\gamma})$) is introduced to reformulate Eq. \eqref{eq:rhopde} as:
\begin{equation}\label{eq:deltapde}
\begin{aligned}
(\prod_1^n\delta^k(\bm{x},\bm{\gamma}))(\mathcal{N}_{\omega^*}[\bm{u}(\bm{x}, t)] - f(\bm{x},t))=0, 
\end{aligned}
\end{equation}
where $\bm{\gamma}$ is geometric parameters to determine the boundary curve function $F_{\bm{\gamma}}(\bm{x})=0$ for $\bm{x}$ at the boundary $\Gamma$, $n$ is the number of patches. 

Compared with DT-PINNs that rely on implicit density ($\rho$) to represent the topology, LT-PINNs introduce the explicit distance function ($\delta$) as topology representation, overcoming the diffuse boundary limitations in DT-PINNs. For systems with multiple topology patches, each patch is characterized by its own geometric distance function, and their collective influence on Eq. \eqref{eq:deltapde} is mediated through the cumulative product of these functions. 

In detail, $\delta(\bm{x}, \bm{\gamma})$ is defined as:
\begin{equation}
\begin{aligned}
\delta(\bm{x}, \bm{\gamma}) &= ~ \frac{1}{1 + e^{-\beta \cdot SDF}}, \\ 
SDF &= \text{sign}(\bm{x}) \cdot d(\bm{x}, F_{\bm{\gamma}}), \\
\text{sign}(\bm{x}) &=
\begin{cases}
-1, & \text{if } x \in \Omega_1,\\
0, & \text{if } x \in \partial \Omega_1,\\
1, & \text{if } x \in \Omega_0,
\end{cases}
\end{aligned}
\label{eq1}
\end{equation}
where $\beta$ is a parameter to determine the sharpness of $\delta$, and $SDF$ is the signed distance function \cite{osher2003constructing}. 

The sign is determined by the relative location of a sample point to the topology. If a sample point is inside the topology ($x \in \Omega_1$), its sign is -1; if it is on the topology boundary ($x \in \partial \Omega_1$), the sign is 0; if it is outside the topology ($x \in \Omega_0$), the sign is 1. $d(\bm{x}, F_{\bm{\gamma}})$ is the distance between a sample point and its nearest boundary point, which can be written as:
\begin{equation}
\begin{aligned}
d(\bm{x}, F_{\bm{\gamma}}) = \|\bm{x}-\bm{x}_{\partial \Omega_1}^*\|_2,
\end{aligned}
\end{equation}
where $\bm{x}_{\partial \Omega_1}^*$ is the nearest boundary point to a given sample point. 

\begin{figure}[tp]%% placement specifier
\centering%% For centre alignment of image.
\includegraphics[width=0.55\textwidth]{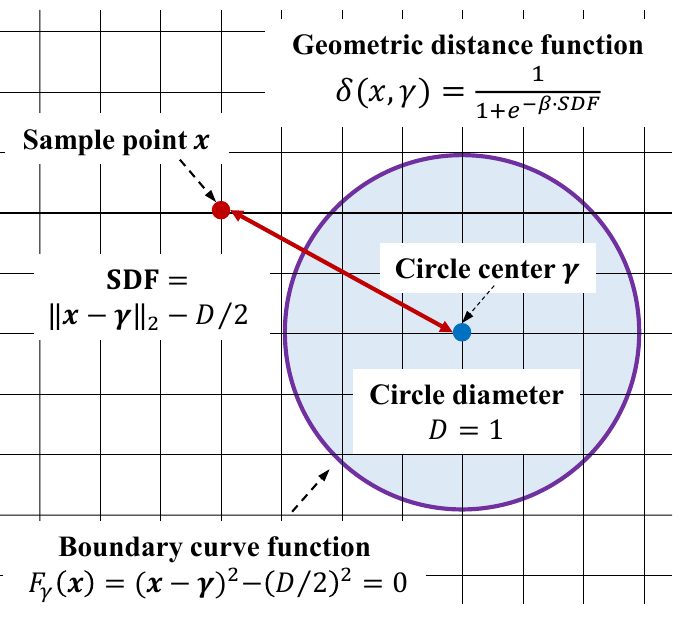}
\caption{An example of geometric distance function with a unit diameter circle topology patch}\label{fig:SDF}
\end{figure}

Taking a circle topology patch with a unit diameter for example (as shown in Fig. \ref{fig:SDF}), its center is denoted by $\bm{\gamma}$. The boundary curve function for this patch is given by $F_{\gamma}(\bm{x})=(\bm{x}-\bm{\gamma})^2-0.5^2=0$, and the corresponding geometric distance function is $\delta(\bm{x}, \bm{\gamma})= \frac{1}{1 + e^{-\beta \cdot (\|\bm{x}-\bm{\gamma}\|_2-0.5)}}$. On the boundary, $\delta=0.5$; inside the patch, $\delta$ decays to 0 quickly; outside the patch, $\delta$ increases to 1 rapidly. It results in a more distinct boundary and a more efficient topology representation compared with implicit density representation used in DT-PINNs.  

In addition to Eq. \eqref{eq:deltapde}, LT-PINNs impose directly boundary condition constraints, as illustrated in Fig. \ref{fig:LTPINN-b}. In general, $\mathcal{B}$ denotes any boundary conditions operator and $g$ is a known function. For Dirichlet boundary conditions, the boundary conditions operator is $\mathcal{B}:\bm{u}=\bm{u}_b^*$. Dirichlet boundary conditions can be enforced without geometric information of the boundary, whereas Neumann boundary conditions depend on the boundary normal vector, represented mathematically as: 
\begin{equation}
\bm{n}=\frac{\nabla F_{\bm{\gamma}}}{\|\nabla F_{\bm{\gamma}}\|_2}, 
\end{equation}
and the Neumann boundary condition operator is:
\begin{equation}
\mathcal{B}:\frac{\partial\bm{u}}{\partial\bm{n}}=\nabla\bm{u} \cdot \frac{\nabla F_{\bm{\gamma}}}{\|\nabla F_{\bm{\gamma}}\|_2}~=q,
\end{equation}
where $q$ is the flux across the boundary. Therefore, the PDE, boundary condition, and data loss functions for LT-PINNs are described as:
\begin{equation}
\begin{aligned}
L_{p} &= \frac{1}{N}\sum^N_{i=1}R_i^2,~ R_i = (\prod_1^n\delta^k(\bm{x},\bm{\gamma}))_i(\mathcal{N}_{\omega^*}[\bm{u}(\bm{x}, t)] - f(\bm{x},t))_i,\\
L_{b}&= 
\begin{cases}
\frac{1}{N_{b}}\sum^{N_{b}}_{i=1}(\bm{u} - \bm{u}_b^*)^2_i, & \text{if Dirichlet b.c.}, \\
\frac{1}{N_{b}}\sum^{N_{b}}_{i=1}(\nabla\bm{u} \cdot \frac{\nabla F_{\bm{\gamma}}}{\|\nabla F_{\bm{\gamma}}\|_2}-q)_i^2, & \text{if Neumann b.c.},\\
\end{cases}
\\
L_{d} &=\frac{1}{N_{d}}\sum^{N_{d}}_{i=1}(\bm{u} - \bm{u}^*)^2_i, 
\end{aligned}
\label{losseq2}
\end{equation}
where $L_{p}$ is the PDE loss, $L_{b}$ is the boundary condition loss, and $L_{d}$ is the data loss, $R$ is the residual of PDE, $\bm{u}_b^*$ are Dirichlet boundary conditions, $\bm{u}^*$ are the sparse measurements/initial conditions on $\bm{u}$, $N$ is the number of sample points for PDE loss, $N_{b}$ is the number of sample points for boundary condition loss, and $N_{d}$ is the number of sample points for data loss.

Since DT-PINNs lack explicit topological representation, they can only enforce Dirichlet boundary conditions indirectly through the composite PDE loss in Eq. \eqref{eq:dirichlet}. This leads to imprecise boundary constraint implementation. Conversely, LT-PINNs employ a dedicated boundary condition loss function (Eq. \eqref{losseq2}) that essentially supports both Dirichlet and Neumann conditions, which also has the potentiality to be applied to arbitrary boundary conditions, yielding precise boundary constraints. This enhanced loss functions enable more accurate PDE solutions and straightforward topology inference without manual interpolation.

Additionally, since most complex topology systems exhibit self-similar patterns and multi-scale hierarchical structures (e.g., lattice-based metamaterials, fractal geometries, or bio-inspired cellular architectures), DT-PINNs' implicit density representation struggles to capture these repetitive features efficiently. In contrast, LT-PINNs' explicit geometric distance functions can naturally encode such periodicity and pattern recurrence through two types of topology loss functions ($L_{t}^1, L_{t}^2$). They are formulated based on the distance between pairs of topology patches, defined as:
\begin{equation}
\begin{aligned}\label{topologyloss}
L_{t}^1 &=\frac{1}{N_{}}\sum^{N_{t}}_{i=1}\sum^{N_{t}}_{j\neq i, j=1}(r_{ij}-r_{ij}^*)^2, \\
L_{t}^2 &=\frac{1}{N_{}}\sum^{N_{}}_{i=1}\sum^{N_{t}}_{j\neq i, j=1}\frac{1}{r_{ij}^2}, 
\end{aligned}
\end{equation}
where $N_{t}$ is the number of topology patches, $r_{ij}$ is the distance between a pair of topology patches, $r_{ij}^*$ is the reference distance, $L_{t}^1$ is the loss related to a fixed distance constraint, and $L_{t}^2$ is the loss related to non-overlapping constraint. Therefore, the total loss function of LT-PINNs that incorporates aforementioned topology loss function is:
\begin{equation}
\begin{aligned}
L_{tot}(\bm{\theta},\bm{\gamma}) &= \lambda_{p} L_{p} + \lambda_{b} L_{b} + \lambda_{d} L_{d} + \lambda_{t} L_{t},  \\
\end{aligned}
\end{equation}
where $L_{tot}$ is the total loss, $\bm{\theta}$ and $\bm{\gamma}$ represent the parameters in the LT-PINN, $\lambda_{t}$ is the weight of $L_{t}$, which is sum of $L_{t}^1$ and $L_{t}^2$, $\lambda_{p}$, $\lambda_{b}$ and $\lambda_{d}$ are the weights of the PDE, boundary condition, and data loss, respectively.

The total loss minimization follows the stochastic gradient descent. In particular, for PDEs with Neumann boundary conditions on a single topology patch, $\bm{\gamma}$ is updated via backpropagation according to the gradient of the corresponding loss: 
\begin{equation}
\begin{aligned}
\bm{\gamma}^{l+1}=&~\bm{\gamma}^l-\eta (\lambda_{p}\frac{\partial L_{p}}{\partial \bm{\gamma}}+\lambda_{b}\frac{\partial L_{b}}{\partial \bm{\gamma}}),\\
\frac{\partial L_{p}}{\partial\bm{\gamma}}&=\frac{1}{N}\sum^N_{i=1}(\prod_1^n\frac{ \partial(\delta^k(\bm{x},\bm{\gamma}))^2}{\partial\bm{\gamma}})_i(\mathcal{N}_{\omega^*}[\bm{u}(\bm{x}, t)] - f(\bm{x},t))^2_i, \\
\frac{\partial L_{b}}{\partial\bm{\gamma}}&=\frac{2}{N_{b}}\sum^{N_{b}}_{i=1}(\nabla\bm{u} \cdot \frac{\nabla F_{\bm{\gamma}}}{\|\nabla F_{\bm{\gamma}}\|_2}-q)_i(\nabla\bm{u} \cdot \frac{\partial\nabla F_{\bm{\gamma}}}{\partial \bm{\gamma}\|\nabla F_{\bm{\gamma}}\|_2})_i,
\end{aligned}
\end{equation}
where $l$ is the training step and $\eta$ is the learning rate. Since $\frac{\partial(\delta^k(\bm{x},\bm{\gamma}))^2}{\partial\bm{\gamma}}$ and $\frac{\partial\nabla F_{\bm{\gamma}}}{\partial \bm{\gamma}\|\nabla F_{\bm{\gamma}}\|_2}$ are rarely zero, it ensures $\bm{\gamma}$ converges to yield solutions fulfilling the PDEs and boundary conditions.

The major difference between LT-PINNs and DT-PINNs lies in the parameterization of $\bm{\gamma}$ as learnable variables in LT-PINNs. This design enables the topology boundary curve function $F_{\bm{\gamma}}(\bm{x})=0$ to dynamically evolve during training, offering three key advantages: (1) elimination of error-prone manual boundary interpolation required in DT-PINNs, (2) precise boundary constraints for both Dirichlet and Neumann boundary conditions, and (3) naturally encoding periodicity and pattern recurrence in complex topology systems. However, LT-PINNs require prior knowledge about $F_{\bm{\gamma}}$.

\section{Numerical Experiments}
\label{sec2}
In this section, we perform a series of numerical experiments to evaluate the accuracy of LT-PINNs by comparing them with DT-PINNs on various PDEs governed by both Dirichlet and Neumann boundary conditions. Moreover, we demonstrate the applicability of LT-PINNs in complex topology optimization problems, involving multi-circle array topology for both steady and unsteady flow scenarios. Finally, we test LT-PINNs on flow rearrangement tasks, considering cases without prior known measurement data. For simplicity, all simulations are restricted to two-dimensional domains and all topology patches are set circles with diameter of 1. Therefore, the boundary curve function of each circle in Eq. \eqref{losseq2} is $F_{\gamma}(\bm{x})=(\bm{x}-\bm{\gamma})^2-0.5^2=0$, where ${\bm{\gamma}}$ means the adaptive center of circle. Consequently, the geometric distance function defined in Eq. \eqref{eq1} in Sec. \ref{subsec:LPINN} is $\delta(\bm{x}, \bm{\gamma})= \frac{1}{1 + e^{-\beta \cdot (\|\bm{x}-\bm{\gamma}\|_2-0.5)}}$, and the normal of the boundary curve is $\bm{n}=2(\bm{x}-\bm{\gamma})$. We choose $\beta=100$ for the geometric distance function to ensure that $\prod_1^n\delta^k(\bm{x},\bm{\gamma})$ has a sharp conversion between 0 and 1. In addition, all trainable parameters $\theta$ in PINNs are initialized by He initialization \cite{he2015delving}, while the $\gamma$ is randomly initialized firstly and then normalized using the sigmoid function to fully span within the ROI. As a result, the optimization of $\gamma$ remains constrained to this predefined ROI spatial domain, avoiding unnecessary computational overhead. 

All the reference solution data is generated by Star-CCM+ 2021.1 and all the training of PINN models are performed on the NVIDIA RTX 4090 GPUs with implementations using the framework PyTorch 2.1.0. The details of the computational setup, e.g., architectures of neural networks, optimizer, and so on, for all test cases are present in \ref{sec:details_computation} in addition to the cases in Sec. \ref{sec:compare}. It should be noted that the computational model setup is tailored for each specific case to achieve optimal results. Data and code are available at: \href{https://github.com/cloud2009/LT-PINN}{\texttt{https://github.com/cloud2009/LT-PINN}}.

\subsection{Benchmarking and Comparative Analysis}
\label{sec:compare}
In this section, we compare DT-PINNs and LT-PINNs in solving for the solution $\bm{u}$ and inferring topology, focusing on the elastic equations with Dirichlet boundary conditions and Laplace's equations with Neumann boundary conditions. Since these two types of boundary conditions serve as the foundation for many others, such as Robin, Cauchy, and symmetry boundary conditions, they are a reasonable choice for test cases. For elastic equations, the Dirichlet boundary conditions, such as the fixed wall boundary, is common in structure analysis. For Laplace's equations, the Neumann boundary conditions, such as constant normal derivative of the solution on the boundary, are also commonly used as illustrative examples. These two types of equations contain second order partial differential terms that are able to represent general features of PDEs.

\begin{figure}[tp]
\centering%% For centre alignment of image.
\subfigure[]{\label{fig:1c_domain-a}
\includegraphics[width=0.65\textwidth]{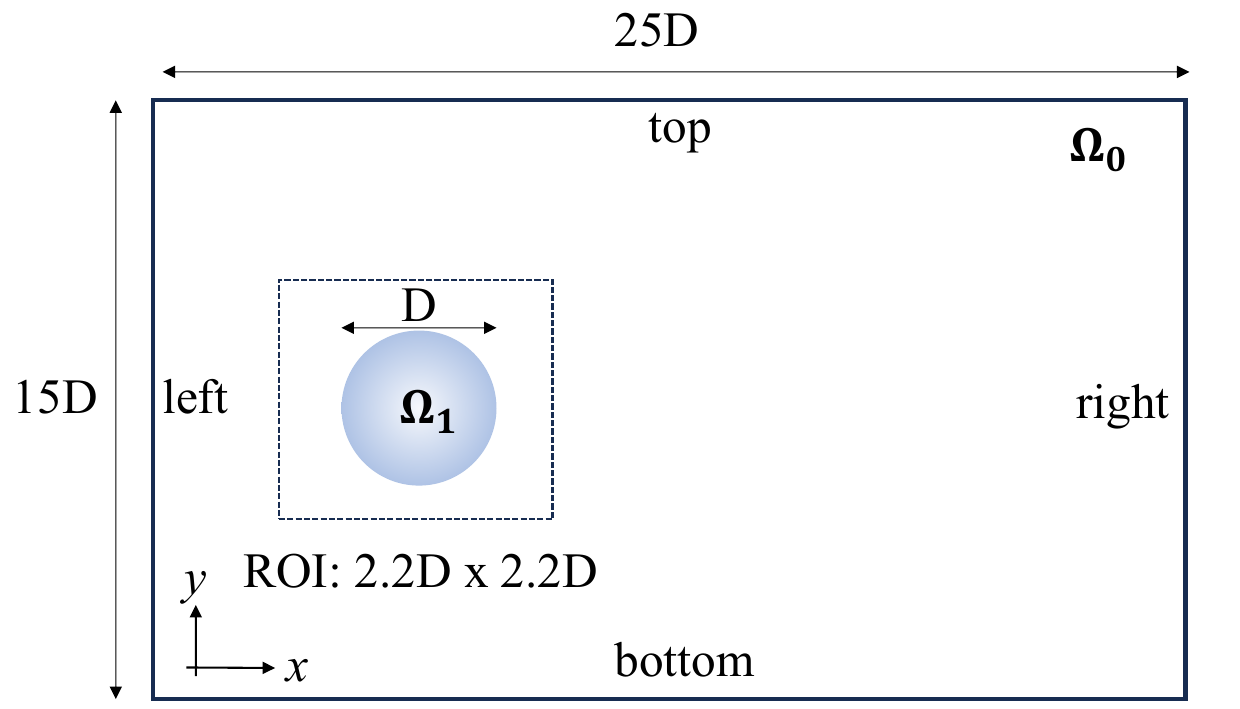}}
\hfill
\subfigure[]{\label{fig:1c_domain-b}
\includegraphics[width=0.6\textwidth]{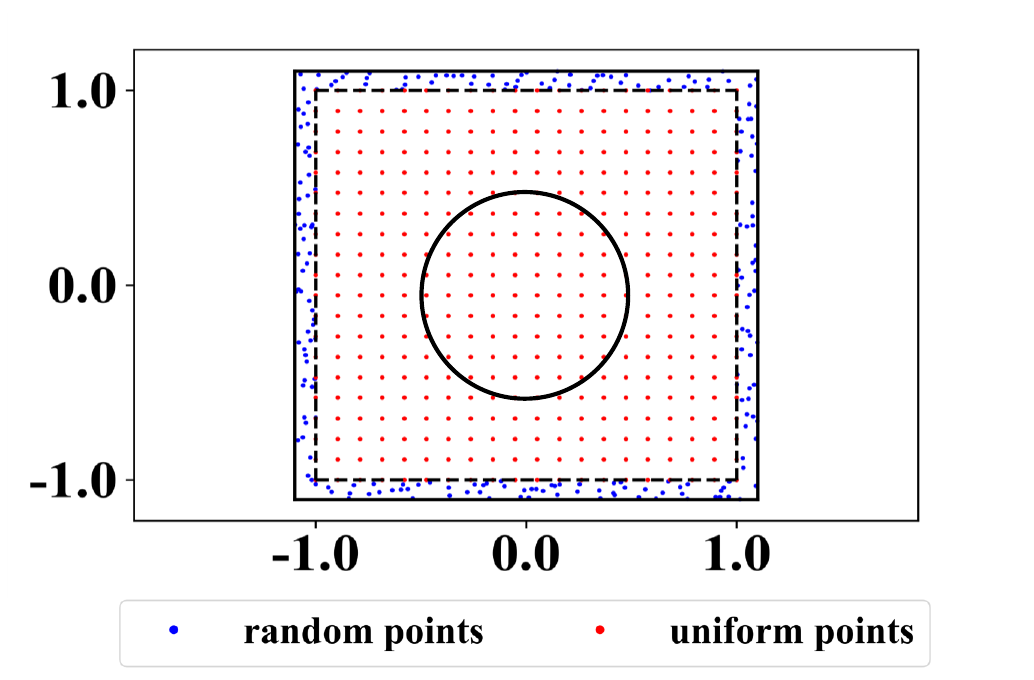}}
\caption{Illustration of computational domain and data sampling strategy: a) the whole simulation domain; b) the data sampling strategy in ROI.}\label{fig:1c_domain}
\end{figure}

The training data for the PINNs are generated within the computational domain illustrated in Fig. \ref{fig:1c_domain-a}. Note that $\Omega_1$ represents the circle topology domain and $\Omega_0$ represents the rest domain for surrounding medium. The diameter of circle is set to 1, i.e., $D=1$. The full domain spans $15D \times 25D$ , while the training data for the PINNs are extracted from a local ROI of $2.2D \times 2.2D$ centered around the circle. Within the ROI, sample points are uniformly distributed inside the $2D \times 2D$ region covering the topological feature, while outside this region, they are randomly distributed, as illustrated in Fig. \ref{fig:1c_domain-b}. Boundary conditions on all four edges of ROI are prescribed according to the generated data. Given the sparse measurement data in the ROI, where the topology is unknown, PINNs are employed to simultaneously determine the solution $\bm{u}$ and identify the topology. 

\subsubsection{Test Case for Elastic Equation with Dirichlet Boundary Conditions}
\label{sec:solid}
We first test the PINNs on the elastic equation with Dirichlet boundary conditions. The elastic equation is defined as:
\begin{equation}\label{eq:solid}
\begin{aligned}
\nabla \cdot \bm{\sigma} ~&=~ 0,\\
\bm{\epsilon}~&=~\frac{1}{2}(\nabla\bm{u}+(\nabla\bm{u})^T),\\
\bm{u} \bigg|_{bc} ~&=~ 0,
\end{aligned}
\end{equation}
where $\bm{\sigma}$ is the normalized stress tensor based on reference stress $\sigma_{ref}$. The three stress components are $[\sigma_{xx},\sigma_{yy},\sigma_{xy}]$. $\bm{\epsilon}$ is the normalized strain tensor based on reference length $u_{ref}$. The three strain components are $[\epsilon_{xx},\epsilon_{yy},\epsilon_{xy}]$. $\bm{u}$ is the normalized displacement based on reference length $u_{ref}$, and it has two components,i.e., $[u_x,u_x]$.

In this study, a linear elastic, isotropic material is considered, which is governed by the following constitutive equation: 
\begin{equation}
\begin{aligned}
\sigma_{xx}~&=~\frac{E}{1-\nu^2}(\epsilon_{xx}+\nu\epsilon_{yy}),\\
\sigma_{yy}~&=~\frac{E}{1-\nu^2}(\epsilon_{yy}+\nu\epsilon_{xx}),\\
\sigma_{xy}~&=~\frac{E}{1-\nu}\epsilon_{xy},\\
\end{aligned}
\end{equation}
where $E$ is Young's module based on the reference strain $\sigma_{ref}$, and $\nu$ is Poisson ratio. The relation between the strain and displacement is:
\begin{equation}
\begin{aligned}
\epsilon_{xx}~&=~\frac{\partial u_x}{\partial x},\\
\epsilon_{yy}~&=~\frac{\partial u_y}{\partial y},\\
\epsilon_{xy}~&=~\frac{1}{2}(\frac{\partial u_x}{\partial y}+\frac{\partial u_y}{\partial x}).\\
\end{aligned}
\end{equation}
To generate data for training, we set $E=1$ and $\nu=0.33$. It is noted that all physical parameters are normalized in this study, to maintain non-dimensional forms, allowing us to focus on the model's general performance. The edge conditions applied on four edges of ROI are specified as follows: free moving (top and bottom edges), fixed (right edge), and uniform pressure of 1 (left edge). The random distributed data points outside region $2D \times 2D$ are used as sparse measurement data, while the uniformly distributed data points inside region $2D \times 2D$ are used to calculate PDE loss. For LT-PINN, the boundary points are uniformly sampled along four concentric rings within the circle topology with different radii at 0.5, 0.4, 0.3, and 0.2. On each concentric ring, the number of sampled boundary points is 128. Detailed description of data preparation and training settings for PINNs are listed in Tab. \ref{tab:solid}.

\begin{table}[ht]
\centering
\begin{tabular}{c c c}
\hline
\textbf{Settings}  & \textbf{DT-PINN}& \textbf{LT-PINN}\\ \hline
\# of layers & 5& 5\\ 
\# of neurons per layer & 64 & 64   \\ 
activation function & Tanh & Tanh   \\ 
optimizer & Adam & Adam  \\ 
learning rate & $1\times10^{-4}$ & $1\times10^{-4}$  \\ 
training epoch & 400,000 & 400,000 \\ 
$N$ & $120\times120$ & $120\times120$ \\ 
$N_{d}$ & 2,167 & 2,167 \\
$N_{b}$ & - & 512 \\
$N_{t}$ & - & - \\
$\lambda_{p}$ & $2\times10^2$& $2\times10^3$   \\ 
$\lambda_{d}$ & $1\times10^4$& $1\times10^4$ \\ 
$\lambda_{b}$ & -& $1\times10^4$  \\ 
$\lambda_{t}$ & - & -   \\ 
PDE &Eq. \eqref{eq:solid} & Eq. \eqref{eq:solid}  \\
\# of topology patches & - &1  \\
\# of GPUs & 1 & 1  \\ \hline
\end{tabular}
\caption{Elastic equation: details of the PINN architectures and operating parameters.}
\label{tab:solid}
% \end{sidewaystable}
\end{table}

\begin{figure}[tp!]
\centering%% For centre alignment of image.
\subfigure[]{\label{fig:solid_field-a}
\includegraphics[width=0.95\textwidth]{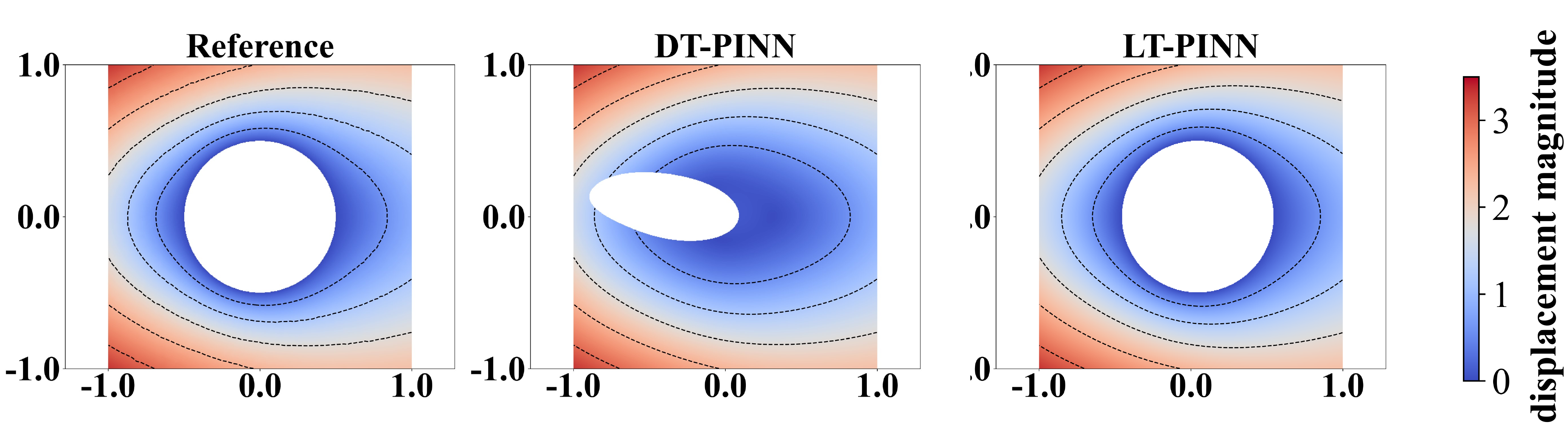}}
\subfigure[]{\label{fig:solid_field-b}
\includegraphics[width=0.95\textwidth]{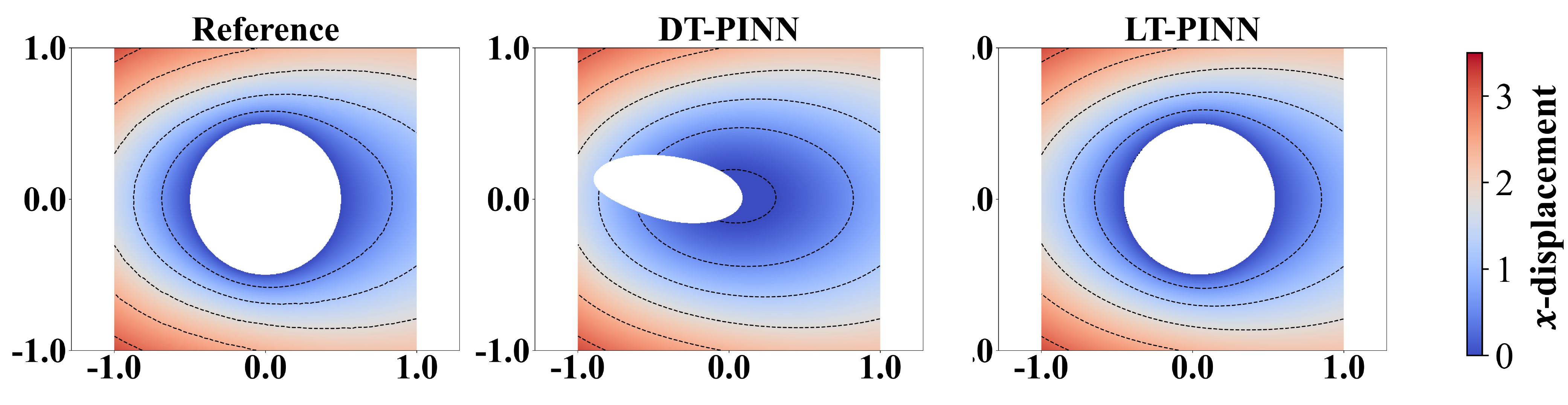}}
\subfigure[]{\label{fig:solid_field-c}
\includegraphics[width=0.95\textwidth]{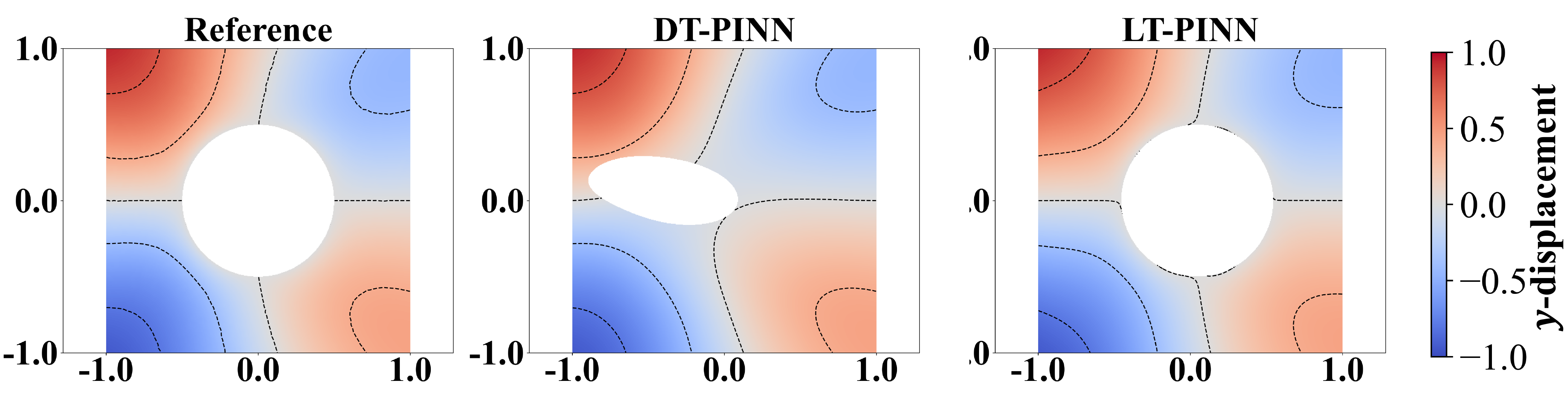}}
\caption{Elastic equation: predicted displacement and topology boundary via different PINNs: a) the displacement magnitude; b) $x$-direction displacement; c) $y$-direction displacement.}\label{fig:solid_field}
\end{figure}

In the DT-PINN, the largest mono-convex topology is obtained by manually thresholding the predicted $\hat \rho$ to define the boundary interface. LT-PINN, however, eliminates this step by directly deriving the topology boundary from the learned curve function $F_{\gamma}(\bm{x})=0$, resulting in an automated and computationally efficient approach. The predictive performance of DT-PINN versus LT-PINN is compared in Fig. \ref{fig:solid_field}, evaluating both the displacement solution $\bm{u}$ and identified topological boundary in the $2D \times 2D$ region. The results demonstrate that LT-PINN accurately locates the circle patch, showing close agreement with the reference topology. In contrast, DT-PINN infers a small elliptical topology, whose center is offset to the left of the reference circle patch. Furthermore, comparison of the displacement fields reveals that LT-PINN achieves better agreement with the reference solution than DT-PINN. Notably, for the small $y$-direction displacement prediction shown in Fig \ref{fig:solid_field-c}, the LT-PINN also predicts more accurate displacement iso-lines than DT-PINN.

\begin{figure}[tp!]
\centering%% For centre alignment of image.
\subfigure[]{\label{fig:solid_field_line-a}
\includegraphics[width=0.95\textwidth]{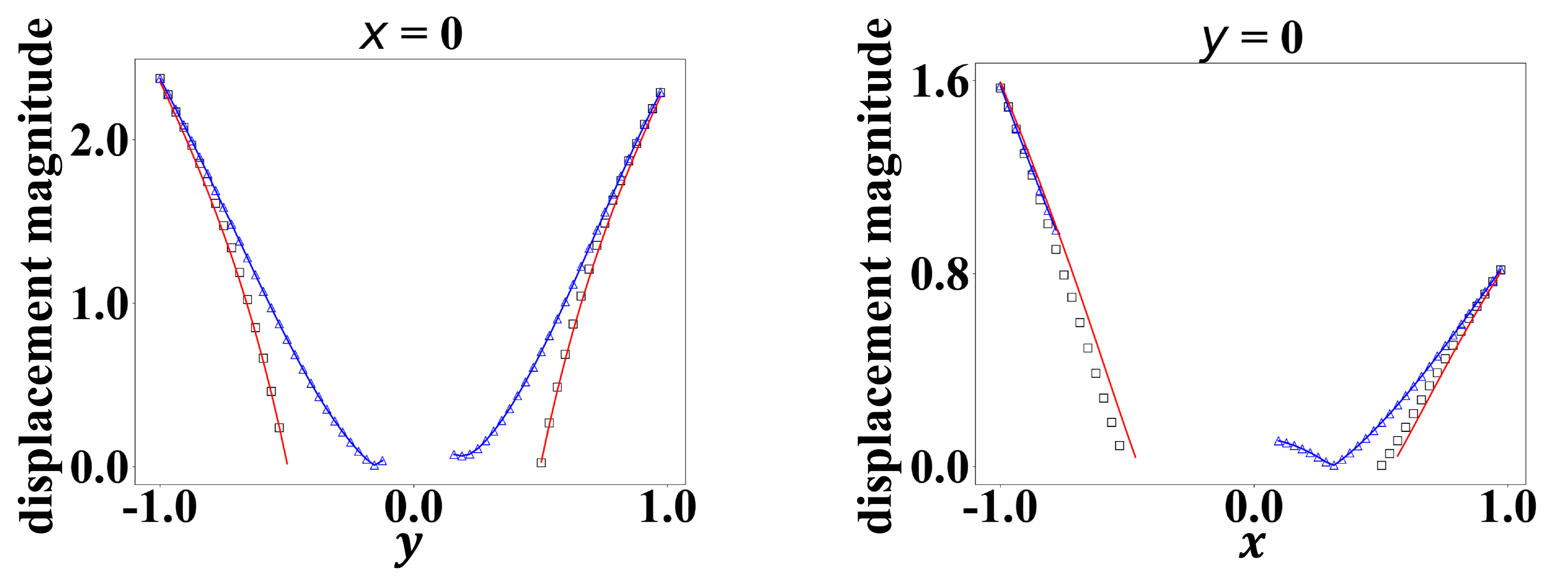}}
\subfigure[]{\label{fig:solid_field_line-b}
\includegraphics[width=0.95\textwidth]{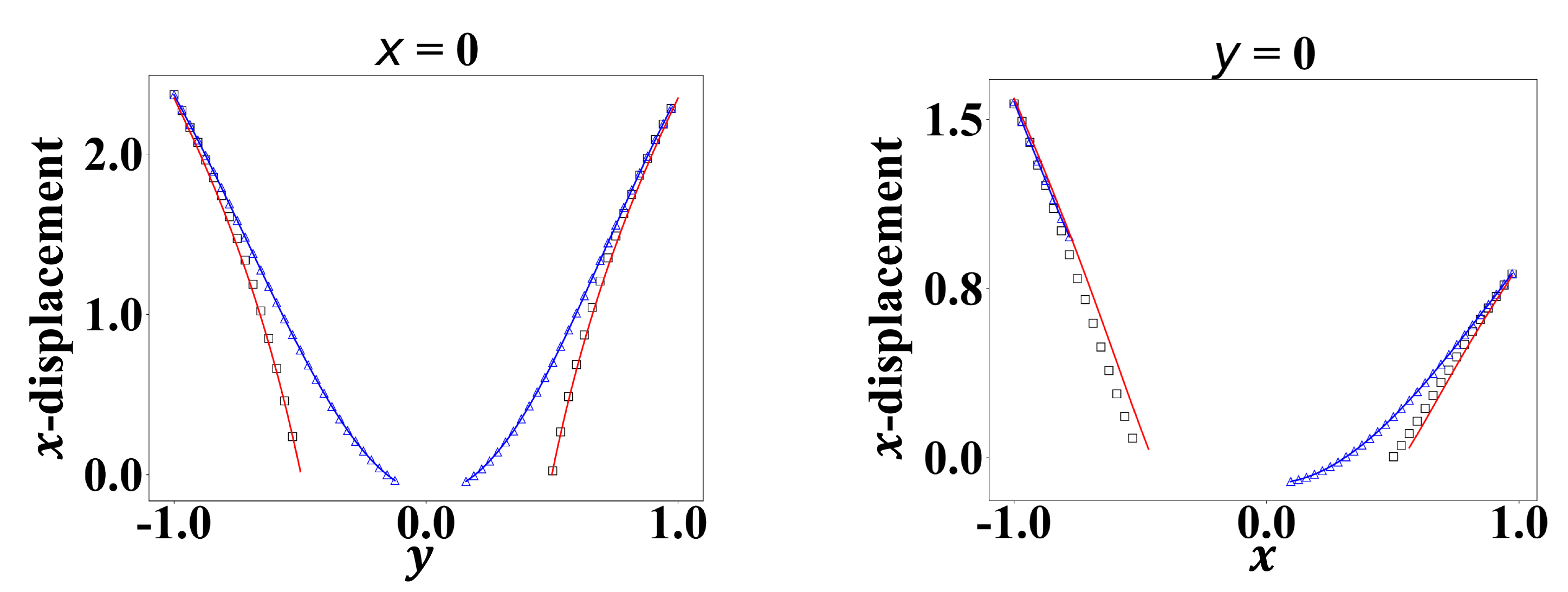}}
\subfigure[]{\label{fig:solid_field_line-c}
\includegraphics[width=0.95\textwidth]{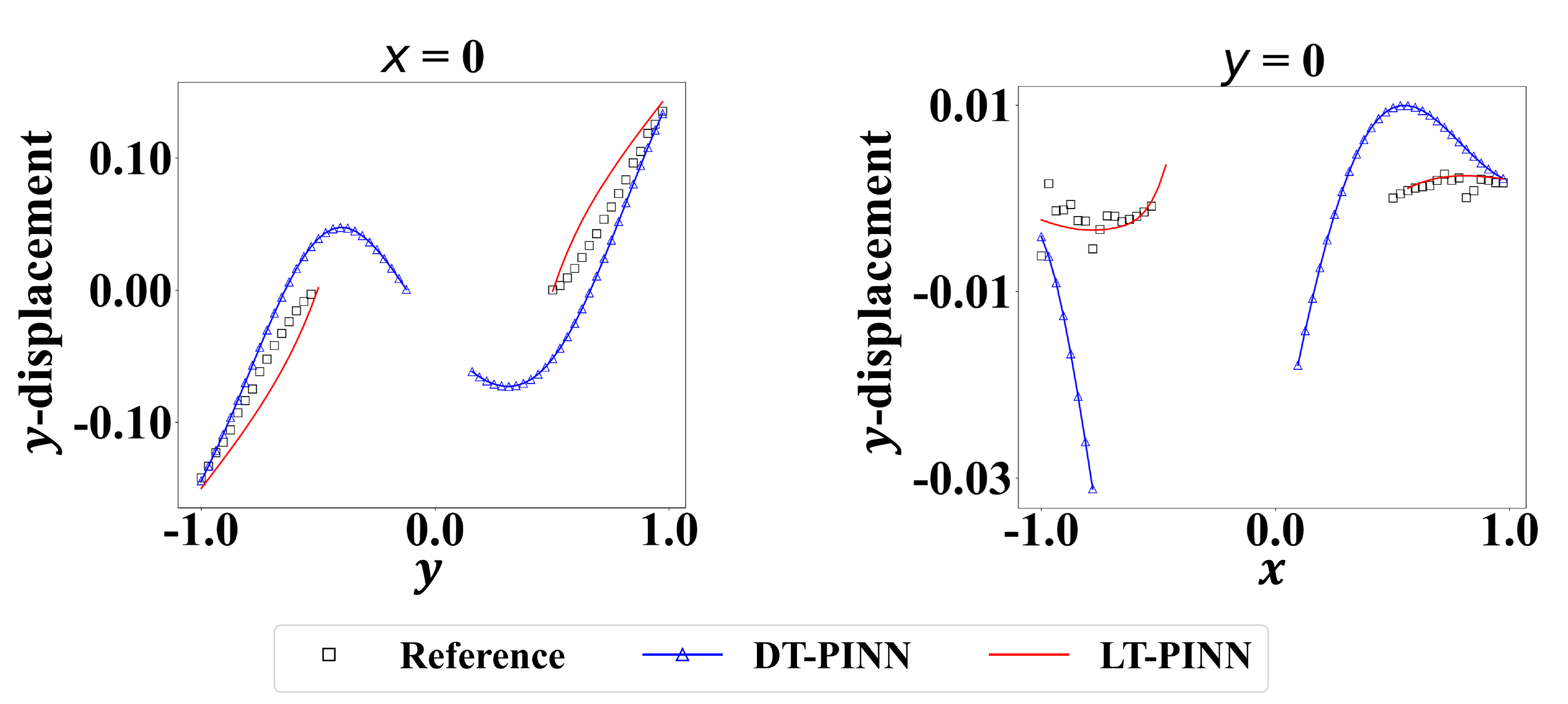}}
\caption{Elastic equation: predicted displacement along the lines $x=0$ and $y=0$: a) the displacement magnitude; b) $x$-direction displacement; c) $y$-direction displacement.}\label{fig:solid_field_line}
\end{figure}

Fig. \ref{fig:solid_field_line} compares the predicted displacements along the lines $x=0$ and $y=0$ with the reference solution. The LT-PINN results demonstrate closer agreement with the reference data for all components ($x$-direction, $y$-direction, and magnitude), consistent with the observations in Fig. \ref{fig:solid_field}. Clearly, while the reference $y$-displacement along $y=0$ exhibits fluctuations due to small numerical interpolation artifacts, LT-PINN produces a physically more plausible smooth profile. These results indicate that LT-PINN not only accurately identifies the unknown topology, but also demonstrates robustness against small numerical artifacts (i.e., interpolation error) in the reference data. 

Moreover, the prediction error is quantified using the relative $L_2$ error, evaluated on a $128 \times 128$ uniform grid of test sample points within the $2D \times 2D$ region (excluding points inside the circle). The corresponding error metrics are presented in Tab. \ref{tab:L2-solid}. LT-PINN demonstrates superior accuracy with significantly lower relative $L_2$ errors compared to DT-PINN. Specifically, the relative $L_2$ errors are reduced by 66.53\% for $x$-direction displacement, 20.46\% for $y$-direction displacement, and 66.43\% for displacement magnitude. These substantial improvements confirm that LT-PINN provides more accurate predictions for both the topological features and displacement fields. Based on the comprehensive analysis presented above, it is convinced that, for topology optimization governed by the elastic equation with Dirichlet boundary conditions, LT-PINN outperforms DT-PINN in both solving the governing PDEs and inferring topology. 

\begin{table}[ht]
\centering
\begin{tabular}{c c c}
\hline
\textbf{Relative $L_2$ error}  & \textbf{DT-PINN}& \textbf{LT-PINN}\\ \hline
$x$-displacement& 0.0998& 0.0334  \\ 
$y$-displacement& 0.0875 & 0.0696  \\ 
Displacement magnitude & 0.0709 & 0.0238  \\ 
\hline
\end{tabular}
\caption{Elastic equation: relative $L_2$ error of predicted displacement via different PINNs.}
\label{tab:L2-solid}
% \end{sidewaystable}
\end{table}

\subsubsection{Test Case for Laplace's Equation with Neumann Boundary Conditions}
\label{sec:heat}
We then use Laplace's equation with Neumann boundary conditions to test PINNs. In this case, the Laplace's equation is: 
\begin{equation} \label{eq:heat}
\begin{aligned}
\nabla^2 T &~= 0, \\
\nabla T \cdot \bm{n}\bigg|_{bc}&~=q,
\end{aligned}
\end{equation}
where $T$ is temperature, $q$ is the heat flux across boundary, and $\bm{n}$ is the normal of boundary. It is noted that both $T$ and $q$ are normalized for simplicity.

To generate data for training, we set $q=-0.5$. The boundary normal's positive direction is defined as pointing the outward from inner circular center. A negative heat flux indicates that the temperature gradient vector opposes the boundary normal direction, implying that the boundary temperature is higher than the exterior. As a result, the circle can be treated as a heat source. The boundary conditions applied on the four edges of ROI are all constant with $T=0$. Training is performed exclusively within the $2D \times 2D$ region, using a $120 \times 120$ uniform grid of sample points for PDE loss evaluation and 1,638 randomly sampled points from this grid as sparse measurement data points of the solution $T$. This formulation treats topology optimization as a high-resolution reconstruction problem from low-resolution measurement data. For LT-PINN, the boundary points are uniformly sampled along four concentric rings within the circle topology with different radii at 0.5, 0.4, 0.3, and 0.2. On each concentric ring, the number of sampled boundary points is 256. Detailed description of data preparation and training settings for PINNs are listed in Tab. \ref{tab:heat}.

\begin{table}[ht]
\centering
\begin{tabular}{c c c}
\hline
\textbf{Settings}  & \textbf{DT-PINN}& \textbf{LT-PINN}\\ \hline
\# of layers & 8& 8\\ 
\# of neurons per layer & 256& 256\\ 
activation function & Tanh & Tanh   \\ 
optimizer & Adam & Adam  \\ 
learning rate & $1\times10^{-4}$ & $1\times10^{-4}$  \\ 
training epoch & 400,000 & 400,000 \\ 
$N$ & $120\times120$ & $120\times120$ \\ 
$N_{d}$ & 1,638& 1,638\\
$N_{b}$ & - & 1,024\\
$N_{t}$ & - & -\\
$\lambda_{p}$ & 1& 1\\ 
$\lambda_{d}$ & $1\times10^4$& $1\times10^4$ \\ 
$\lambda_{b}$ & -& 1\\ 
$\lambda_{t}$ & - & -   \\ 
PDE &Eq. \eqref{eq:heat} & Eq. \eqref{eq:heat}  \\
\# of topology patches & - &1  \\
\# of GPUs & 1 & 1  \\ \hline
\end{tabular}
\caption{Laplace's equation: details of the PINN architectures and operating parameters.}
\label{tab:heat}
% \end{sidewaystable}
\end{table}

\begin{figure}[tp]
\centering%% For centre alignment of image.
\includegraphics[width=0.98\textwidth]{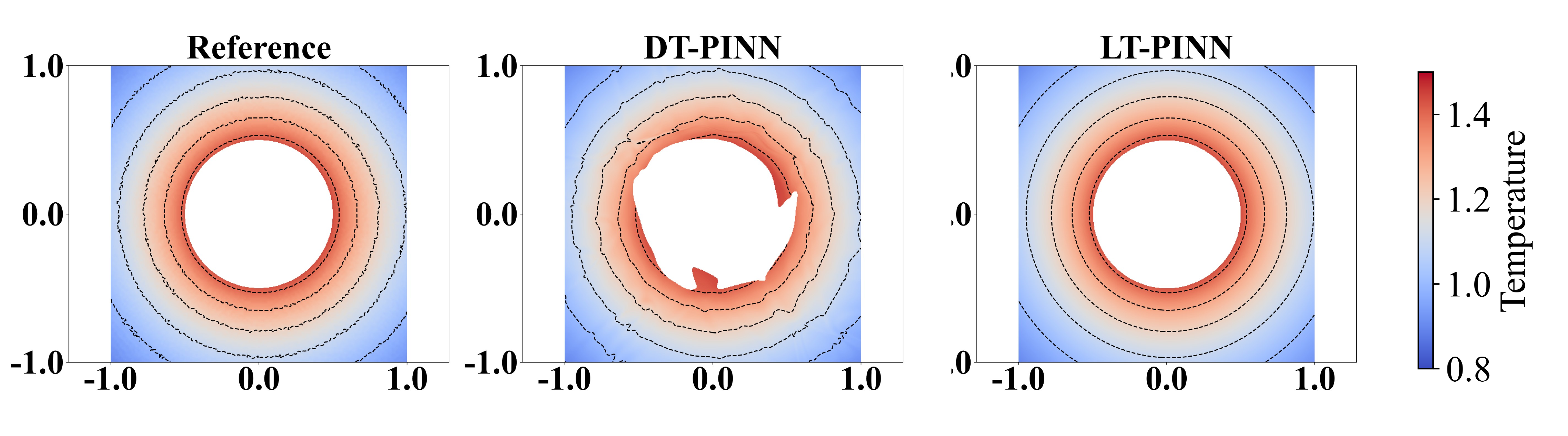}
\caption{Laplace's equation: reconstructed temperature field and topology boundary via different PINNs.}\label{fig:heat_field}
\end{figure}

As $\hat\rho$ is the density representation of topology in DT-PINNs (see Eq. \eqref{eq:dirichlet}), a threshold value of the predicted $\hat\rho$ is needed to separate topology from its surrounding domain. It is manually selected to ensure the largest mono-convex topology is distinguished in DT-PINNs. However, for LT-PINNs, it does not require manual interpolation to extract topology. In LT-PINNs, the optimized parameter $\gamma$ and the boundary function $F{\gamma}(\bm{x})=0$ explicitly define the topology, offering greater reliability and efficiency compared to DT-PINNs. The resulting reconstructed temperature $T$ and topology boundary in region $2D \times 2D$ are shown in Fig \ref{fig:heat_field}. The results demonstrate that LT-PINN accurately identifies both the position and shape of the circle, showing close agreement with the reference solution. In contrast, DT-PINN produces an irregular topology with prominent spikes. Furthermore, temperature field reconstruction reveals LT-PINN's enhanced prediction accuracy. DT-PINN, on the other hand, generates less smooth temperature field characterized by circumferential fluctuations in the iso-lines, which is directly correlated with DT-PINN's irregular topology boundary. 

\begin{figure}[ht!]
\centering%% For centre alignment of image.
\subfigure[]{\label{fig:heat_field_line-a}
\includegraphics[width=0.88\textwidth]{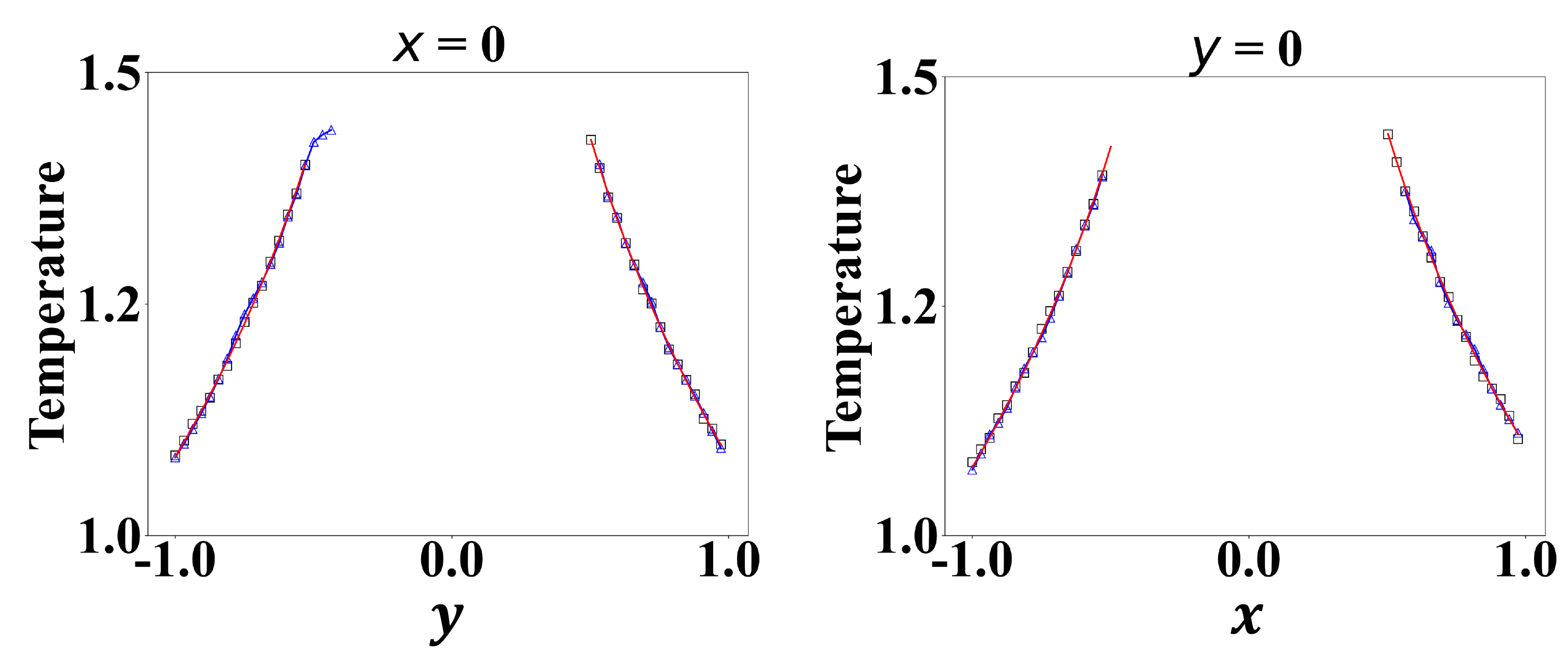}}
\subfigure[]{\label{fig:heat_field_line-b}
\includegraphics[width=0.88\textwidth]{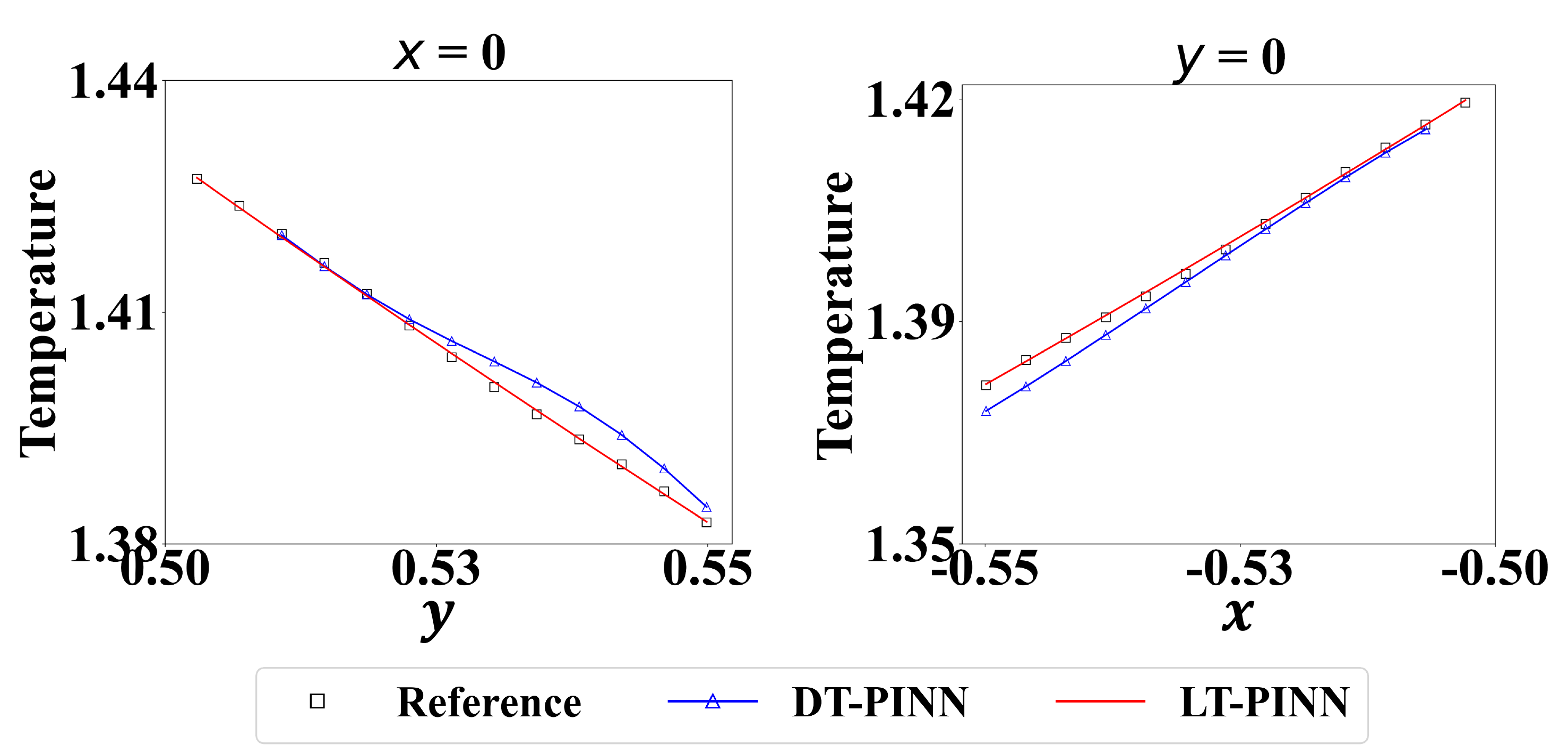}}
\caption{Laplace's equation: reconstructed temperature along $x=0$ and $y=0$ lines: a) temperature along the whole line; b) zoom-in view of temperature near boundary.}\label{fig:heat_field_line}
\end{figure}

The temperature profiles along two lines $x=0$ and $y=0$ are also analyzed, with zoom-in views of the boundary regions extracted and shown in Fig. \ref{fig:heat_field_line}. The temperature profiles along both lines show minimal discrepancies between the reference data and both models' predictions, given low-resolution reference temperature as measurement data for training. A minor discrepancy of the temperature profile along the line $x=0$ (Fig. \ref{fig:heat_field_line}a) from DT-PINN is observed around $y=-0.5$, due to the irregularly reconstructed topology. However, a closer examination of the boundary regions (Fig. \ref{fig:heat_field_line}b) reveals noticeable deviations between the reference solution and DT-PINN's prediction, while LT-PINN's reconstructed temperature profiles maintain better agreement with the reference data. 

\begin{figure}[ht]
\centering%% For centre alignment of image.
\includegraphics[width=0.6\textwidth]{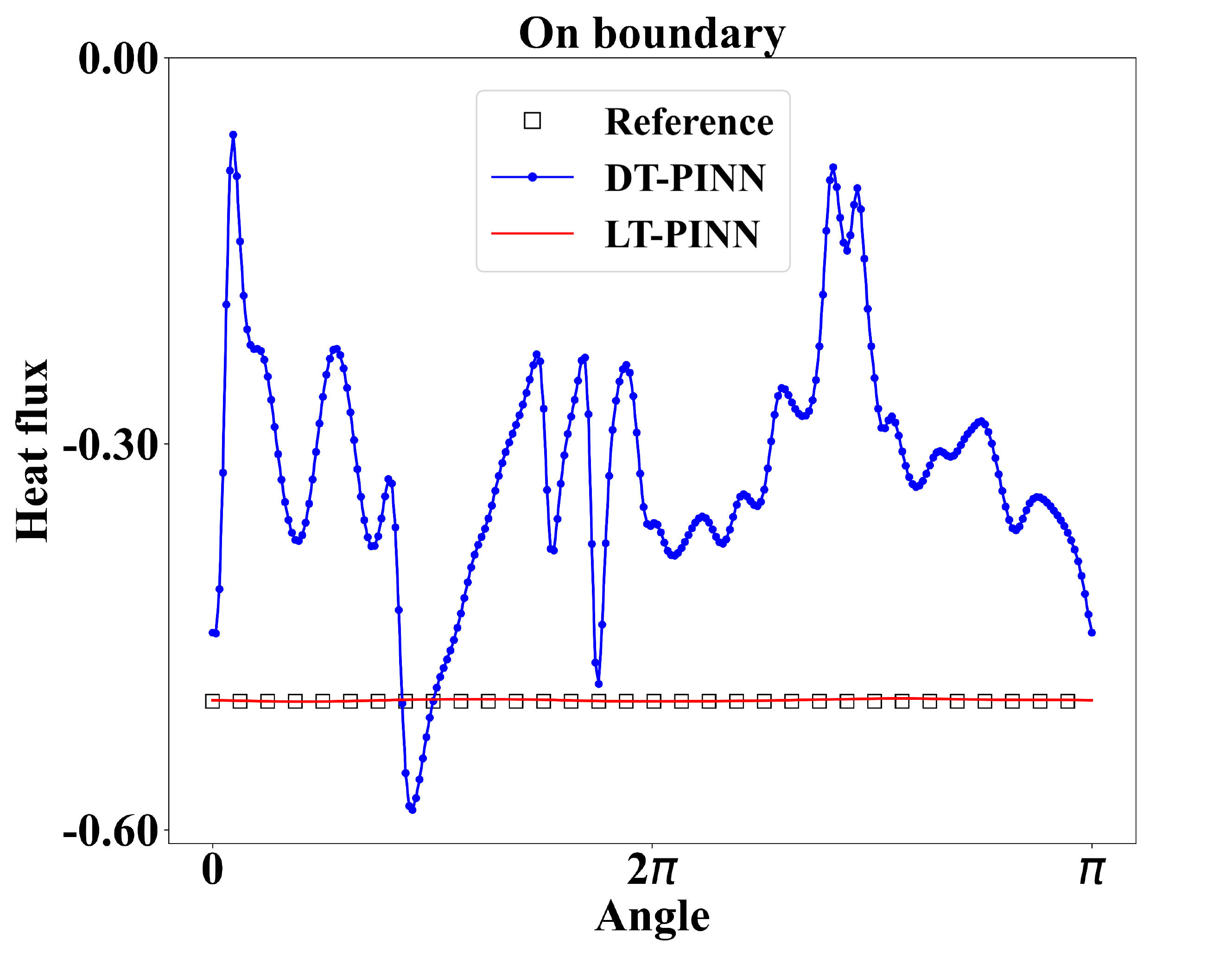}
\caption{Laplace's equation: reconstructed heat flux across the reference boundary via different PINNs.}\label{fig:heatflux}
\end{figure}

The heat flux (i.e., normal gradient of temperature) across the reference boundary is also present in Fig. \ref{fig:heatflux}. Obviously, DT-PINN fails to predict a constant heat flux across the boundary defined by the Neumann boundary conditions (i.e., $q=-0.5$), showing significant deviations from the reference. It tends to predict a smaller and random heat flux, i.e., a smaller temperature gradient on the boundary, which is supposed to be a sign of spectral bias of PINNs \cite{rahaman2019spectral}, \cite{xu2024overview}. Conversely, LT-PINN predicts almost the same constant heat flux that closely matches the reference.

The accuracy of the reconstructed temperature field is quantified using the relative $L_2$ error, evaluated on a $128 \times 128$ uniform grid of test sample points within the $2D \times 2D$ domain (excluding points inside the circle). Moreover, the error in the heat flux across the boundary is computed. Both error metrics are summarized in Tab. \ref{tab:L2-heat}. Despite using identical low-resolution temperature data in the data loss function for both methods, LT-PINN achieves significantly lower errors: a 44.90\% reduction in the relative $L_2$ error for the temperature field and a 99.42\% reduction for the boundary heat flux, compared to DT-PINN. This disparity arises because DT-PINN fails to enforce Neumann boundary conditions in its total loss function, resulting in inaccurate boundary predictions and degraded temperature field reconstruction. Conversely, LT-PINN explicitly incorporates these boundary conditions, leading to improved accuracy. Therefore, for topology optimization governed by Laplace's equation with Neumann boundary conditions, LT-PINN excels DT-PINN in predictive accuracy for both the PDE solution and topology. This advantage stems primarily from LT-PINN's rigorous incorporation of Neumann boundary conditions into its total loss function. 

\begin{table}[ht]
\centering
\begin{tabular}{c c c}
\hline
\textbf{Relative $L_2$ error}  & \textbf{DT-PINN}& \textbf{LT-PINN}\\ \hline
Temperature & 0.0049 & 0.0027  \\ 
Heat flux across boundary & 0.4006 & 0.0023  \\ 
\hline
\end{tabular}
\caption{Laplace's equation: relative $L_2$ error of reconstructed temperature and heat flux via different PINNs.}
\label{tab:L2-heat}
% \end{sidewaystable}
\end{table}

\subsection{Exploring LT-PINNs for Complex Topology Optimization}
\begin{figure}[ht]
\centering%% For centre alignment of image.
\subfigure[]{\label{fig:nDomain-a}
\includegraphics[width=0.3\textwidth]{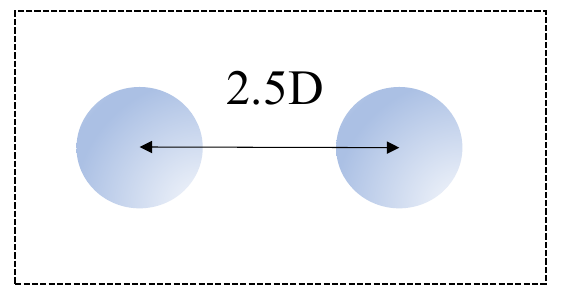}}
\hfil
\subfigure[]{\label{fig:nDomain-b}
\includegraphics[width=0.25\textwidth]{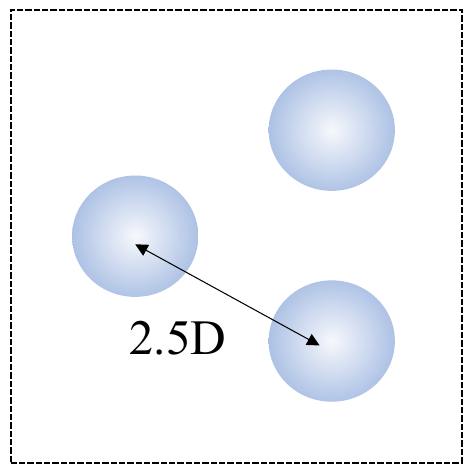}}
\hfil
\subfigure[]{\label{fig:nDomain-c}
\includegraphics[width=0.4\textwidth]{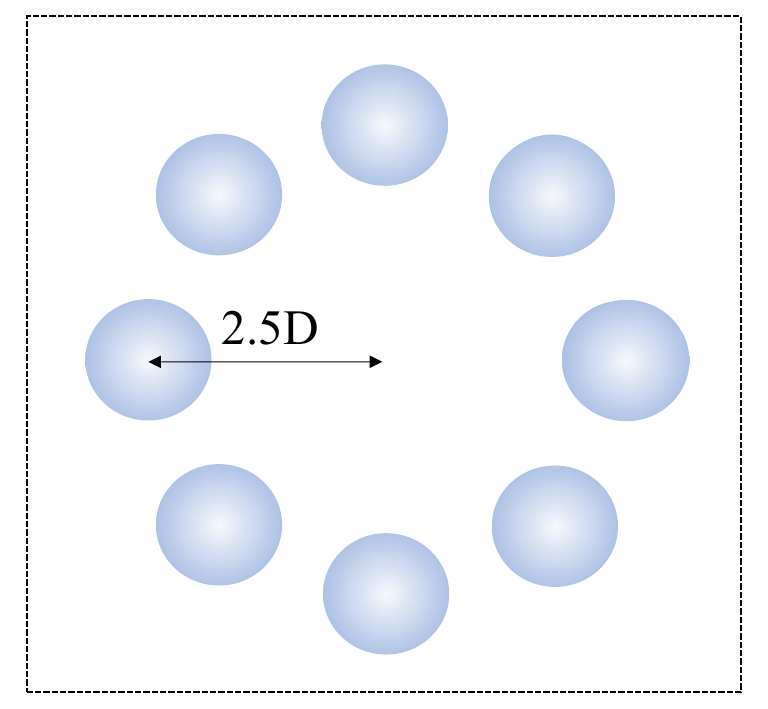}}
\caption{Time-dependent flow: configurations of multi-circle arrays: a) 2-circle array; b) 3-circle array; c) 8-circle array.}\label{fig:nDomain}
\end{figure}

The strengths and capabilities of LT-PINNs have been thoroughly validated in Sec. \ref{sec:compare} through comparative benchmarking against DT-PINNs. Given these promising results, we now explore LT-PINN’s potential in tackling more complex problems, further evaluating its topology optimization performance and robustness. In this section, optimization of a complex topology represented by a multi-circle array, governed by highly nonlinear Navier-Stokes equations, is carried out. The entire simulation domain is $25D\times15D$, similar to that in Fig. \ref{fig:1c_domain}a, and the number of circles in the domain to be analyzed is set 2, 3, and 8, with the respective configurations shown in Fig. \ref{fig:nDomain}. For the 2-circle array, the center-to-center distance between the two circles is fixed at $2.5D$. In the 3-circle array, each pair of adjacent circles maintains the same $2.5D$ spacing. For the 8-circle array, the circles are uniformly distributed on a circumference of the $2.5D$-radius.

In Sec. \ref{sec:compare}, only single circle topology is considered in the test, thus it does not require topology loss function that is related to the constraints of multiple circles in Eq. \eqref{topologyloss}. However, for multi-circle array topologies with self-similar patterns, we include the topology loss function in the LT-PINN besides the other loss functions, as topology loss function can reduce unnecessary optimization space and enhance LT-PINN's performance dealing with complex topology. In detail, for 2-circle array and 3-circle array, we consider the fixed distance constraint loss function ($L_{t}^1$), that forces distance between each pair of circles to be $2.5D$, as shown in Fig. \ref{fig:nDomain-a} and \ref{fig:nDomain-b}. For 8-circle array, the fixed distance constraint loss function ($L_{t}^1$) forces distance between any circle to the center of 8-circle array to be $2.5D$, as shown in Fig. \ref{fig:nDomain-c}. In addition to $L_{t}^1$, we add the non-overlapping constraint loss function ($L_{t}^2$) between each pair of circles for 8-circle array, which aims to maximize their pairwise distance so as to avoid overlapping.

The ROI for model training is adaptively adjusted to the dimensions of the multi-circle array, with a 1.1$D$ extension in both width and height to ensure full coverage of all circles. The dimensions are described as: $[L_{x,min}-1.1D,L_{x,max}+1.1D]\times[L_{y,min}-1.1D,L_{y,max}+1.1D]$, where the minimal and maximum positions of the circle center in $x,y$ directions can be defined by $L_{x,min},L_{x,max},L_{y,min},L_{y,max}$. 

\subsubsection{Time-Independent Flow Problem}
\label{sec:steady}
 The flow problem governed by the time-independent Navier-Stokes equation is first investigated. The time-independent Navier-Stokes equation is given below:
\begin{equation}\label{eq:steadyNS}
\begin{aligned}
\nabla \cdot \bm{u} &~= 0, \\
(\bm{u} \cdot \nabla)\bm{u} &~= -\nabla{p} + \frac{1}{Re}\Delta{\bm{u}},\\
\bm{u} \bigg|_{bc} &= 0,
\end{aligned}
\end{equation}
where $\bm{u}$ is the velocity vector, $p$ is the pressure, and $Re$ is the Reynolds number. A no-slip boundary condition is imposed on all fixed circles. 

Since it is a time-independent flow problem, we choose Reynolds number $Re=1$ for laminar flow in Eq. \eqref{eq:steadyNS}. To generate training data, the edge conditions of ROI are: symmetry conditions on the top and bottom edges, zero pressure on the right edge, and uniform unit velocity on the left edge. Given the solution data from the numerical simulation, we extract sampled points for model training, with sparse measurement data being randomly selected from points outside the core part of ROI, i.e.,$[L_{x,min}-D,L_{x,max}+D]\times[L_{y,min}-D,L_{y,max}+D]$. Inside the core part of ROI, uniformly distributed PDE points are extracted for evaluating PDE loss in Eq. \eqref{losseq2}. Furthermore, to enforce the no-slip boundary conditions represented by Dirichlet boundary conditions on each circle, we uniformly sample 128 boundary points along four concentric rings within each circle at radii of 0.2, 0.3, 0.4, and 0.5. The velocity at all sampled points is set to zero to satisfy the no-slip condition. Detailed description of the data generation and training settings for LT-PINNs are listed in Tab. \ref{tab:apend}. 

\begin{figure}[tp]
\centering
\begin{tabular}{cc}
\begin{minipage}[b]{0.35\textwidth}
    \centering
    \subfigure{\label{fig:1d2c-a}
    \includegraphics[width=\textwidth]{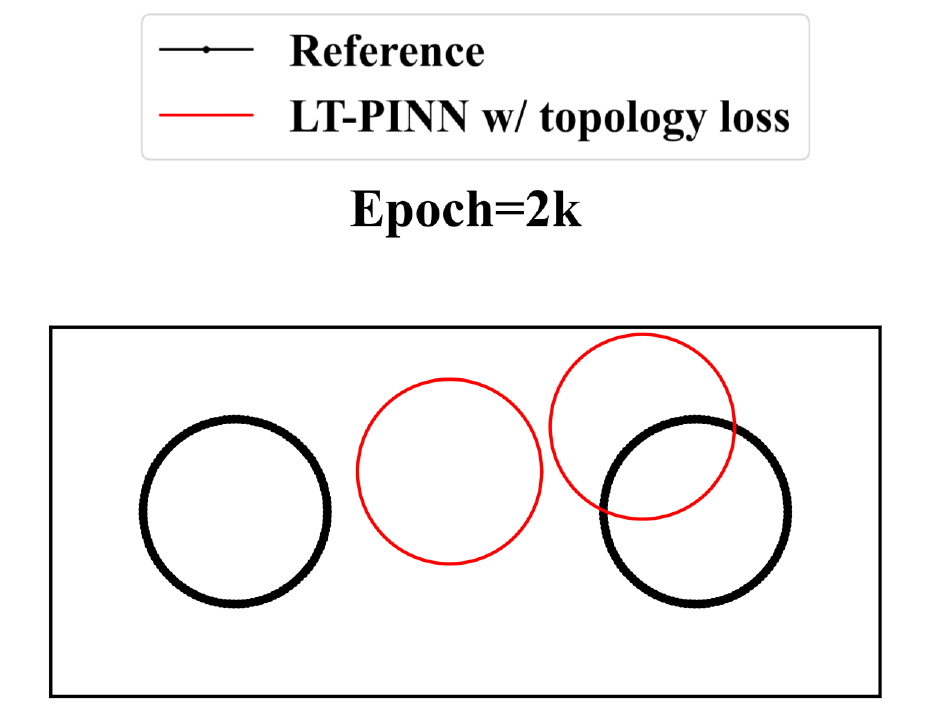}}
    (a)
\end{minipage}
&
\begin{minipage}[b]{0.35\textwidth}
    \centering
    \subfigure{\label{fig:1d2c-b}
    \includegraphics[width=\textwidth]{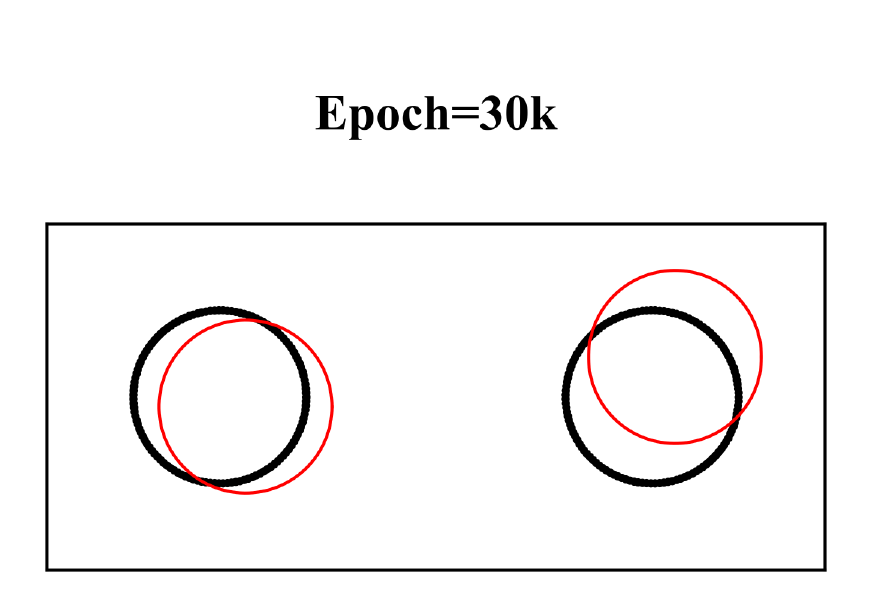}}
    (b)
\end{minipage}
\\
\multirow{2}{*}{
\begin{minipage}[t]{0.455\textwidth}
    \centering
    \subfigure{\label{fig:1d2c-e}
    \includegraphics[width=\textwidth]{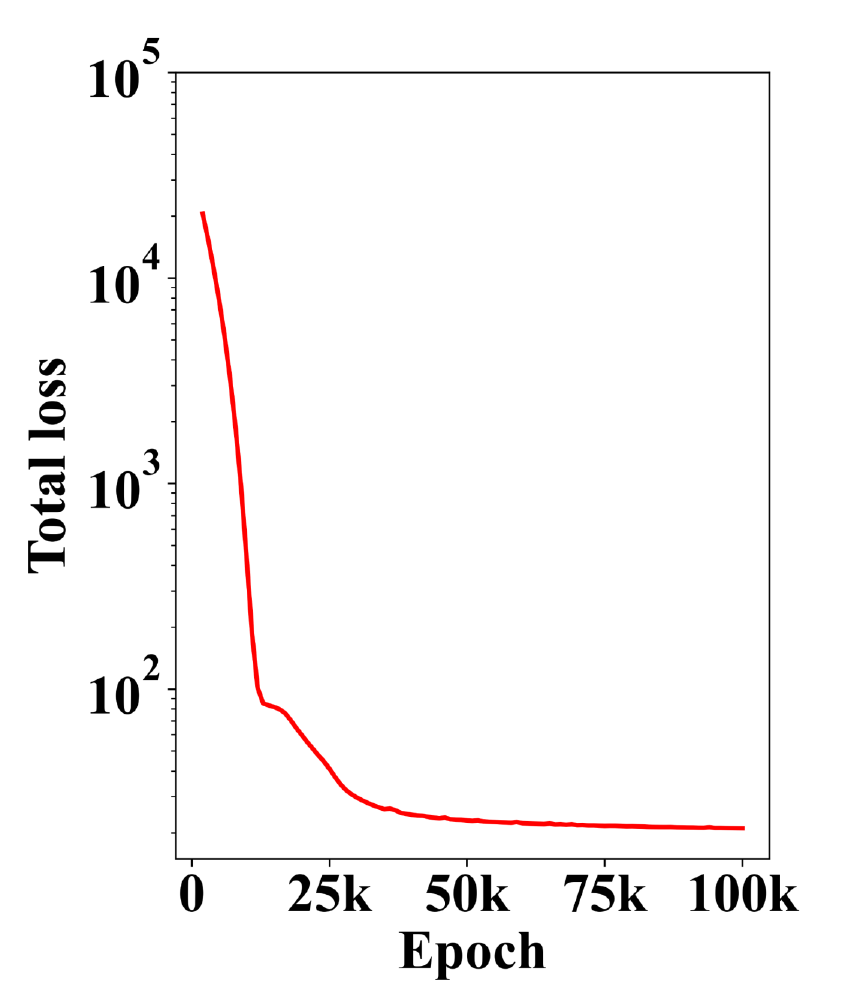}}
    (e)
\end{minipage}}
&
\begin{minipage}[t]{0.35\textwidth}
    \centering
    \subfigure{\label{fig:1d2c-c}
    \includegraphics[width=\textwidth]{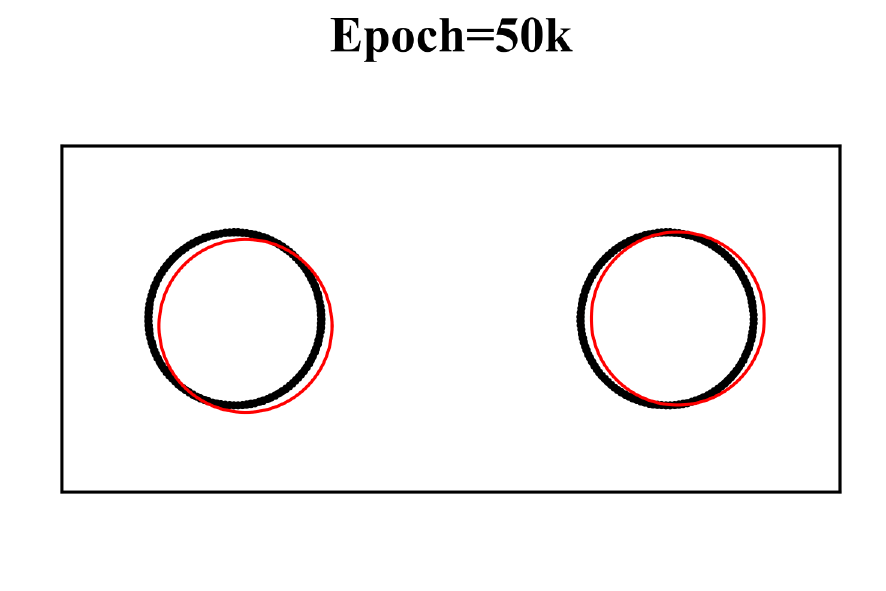}}
    (c)
\end{minipage} 
\\
&
\begin{minipage}[t]{0.35\textwidth}
    \centering
    \subfigure{\label{fig:1d2c-d}
    \includegraphics[width=\textwidth]{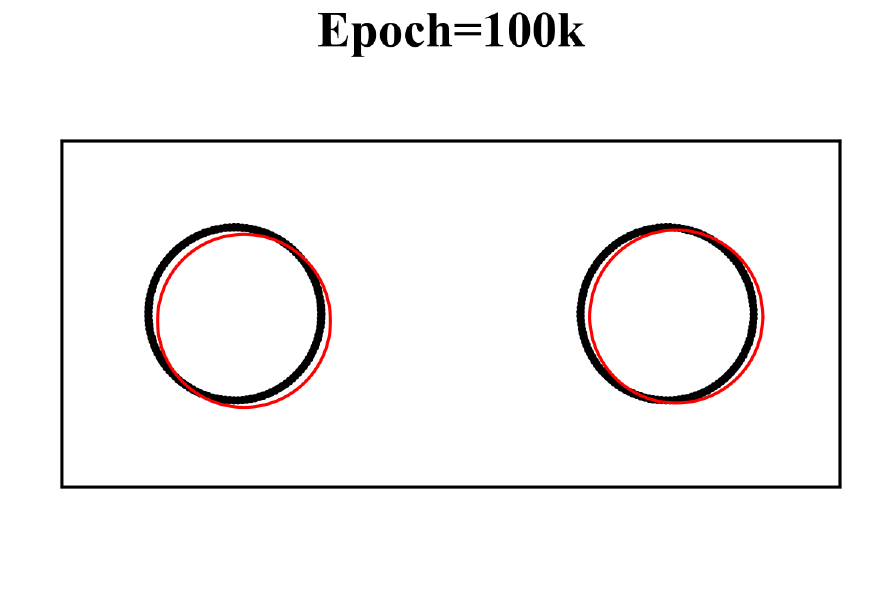}}
    (d)
\end{minipage} 
\end{tabular}
\caption{Time-independent flow: loss history and $\gamma$ trajectory for 2-circle array: a-d) $\gamma$ trajectory at different training epochs; e) loss history.}
\label{fig:1d2c}
\end{figure}

Initially, $\gamma$ (representing the circle centers) is randomly distributed within the domain. During training, the predicted topology of the multi-circle array, determined by $\gamma$, gradually converges toward the reference configuration. The training loss history and $\gamma$ trajectory of the 2-circle array is shown in Fig. \ref{fig:1d2c}. As the training loss diminishes, $\gamma$ progressively aligns with the reference. When the loss becomes stable after around 50k epochs, $\gamma$ also tends to be stable. Consistent findings on the loss history and $\gamma$ trajectory are also observed for the 3-circle array and the 8-circle array. Hence, we refrain from further elaboration. LT-PINNs enable simultaneous flow field prediction and topology optimization. Clearly, since $\gamma$ can be monitored concurrently with the loss history, we find that an early stopping strategy also becomes viable once $\gamma$ stabilizes, significantly reducing training time. 

\begin{figure}[ht]
\centering%% For centre alignment of image.
\subfigure[]{\label{fig:1d2c_field-a}
\includegraphics[width=0.8\textwidth]{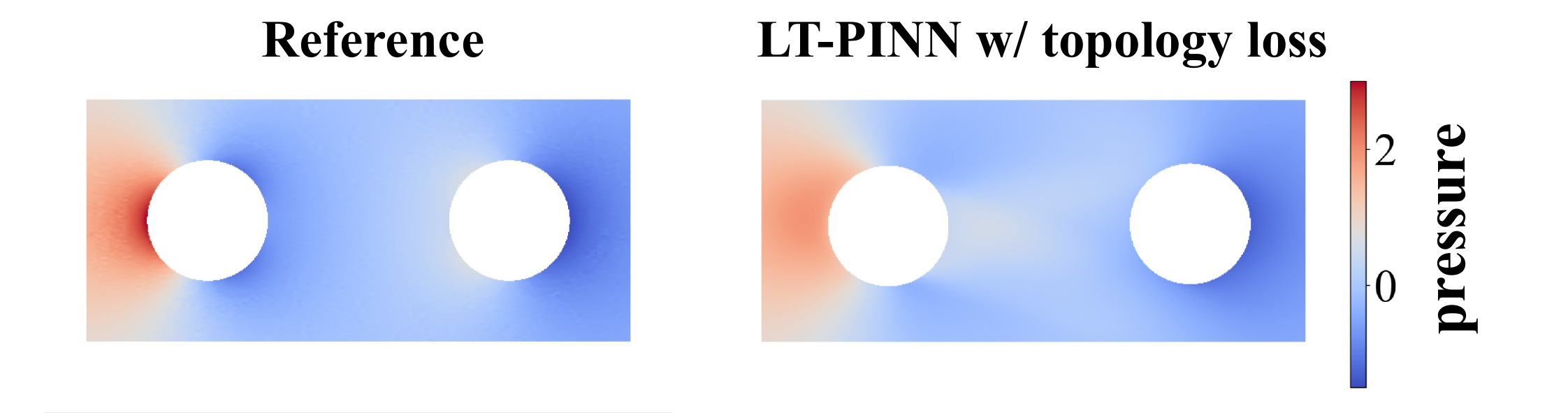}}
\subfigure[]{\label{fig:1d2c_field-b}
\includegraphics[width=0.8\textwidth]{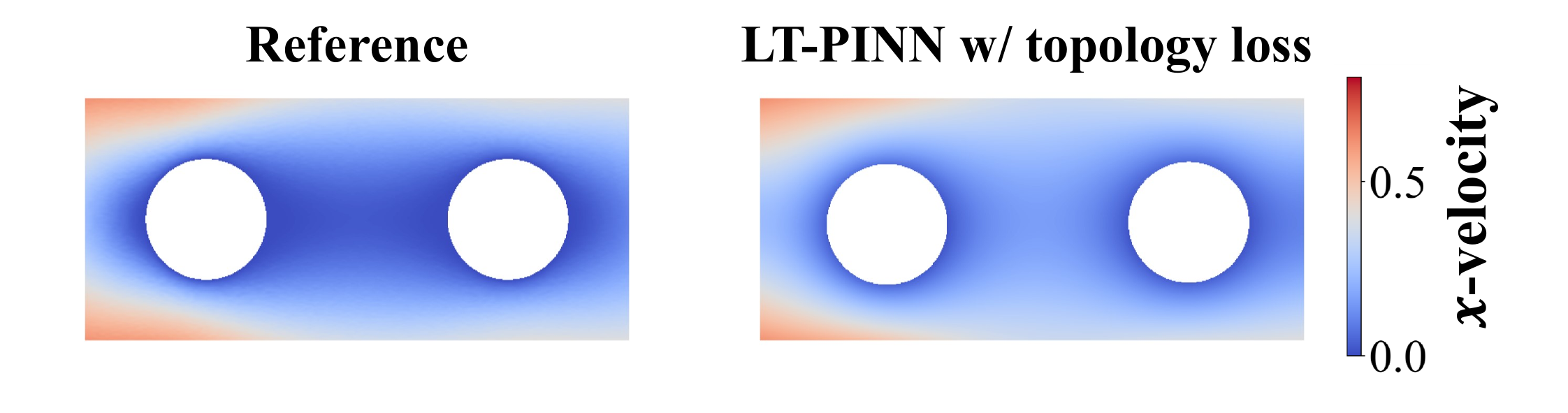}}
\subfigure[]{\label{fig:1d2c_field-c}
\includegraphics[width=0.8\textwidth]{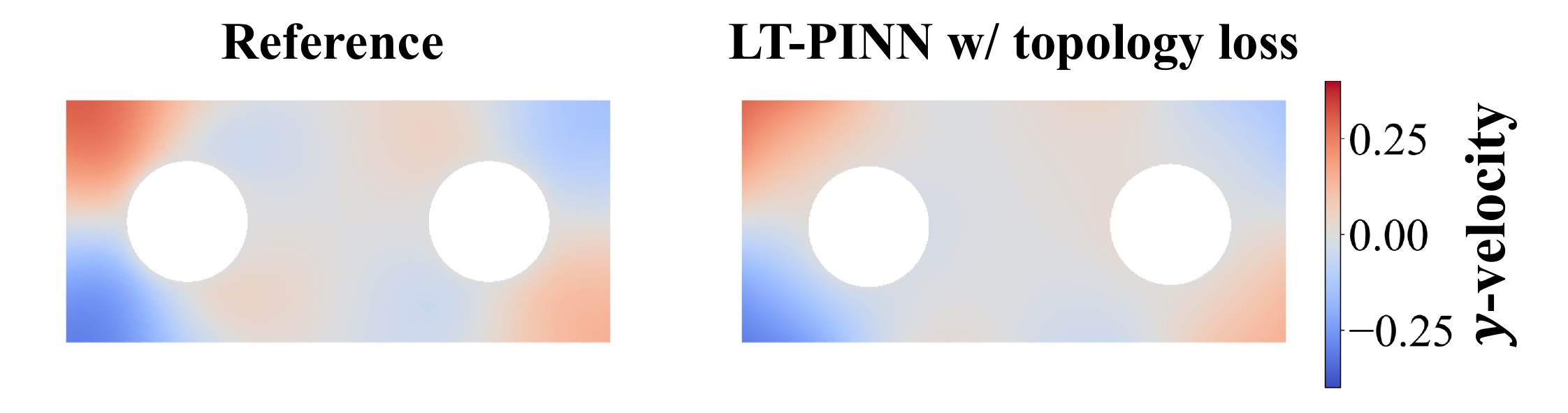}}
\caption{Time-independent flow: predicted velocity and pressure field via LT-PINN for 2-circle array: a) pressure; b) $x$-velocity; c) $y$-velocity.}\label{fig:1d2c_field}
\end{figure}

\begin{figure}[tp]
\centering%% For centre alignment of image.
\subfigure[]{\label{fig:1d3c_field-a}
\includegraphics[width=0.7\textwidth]{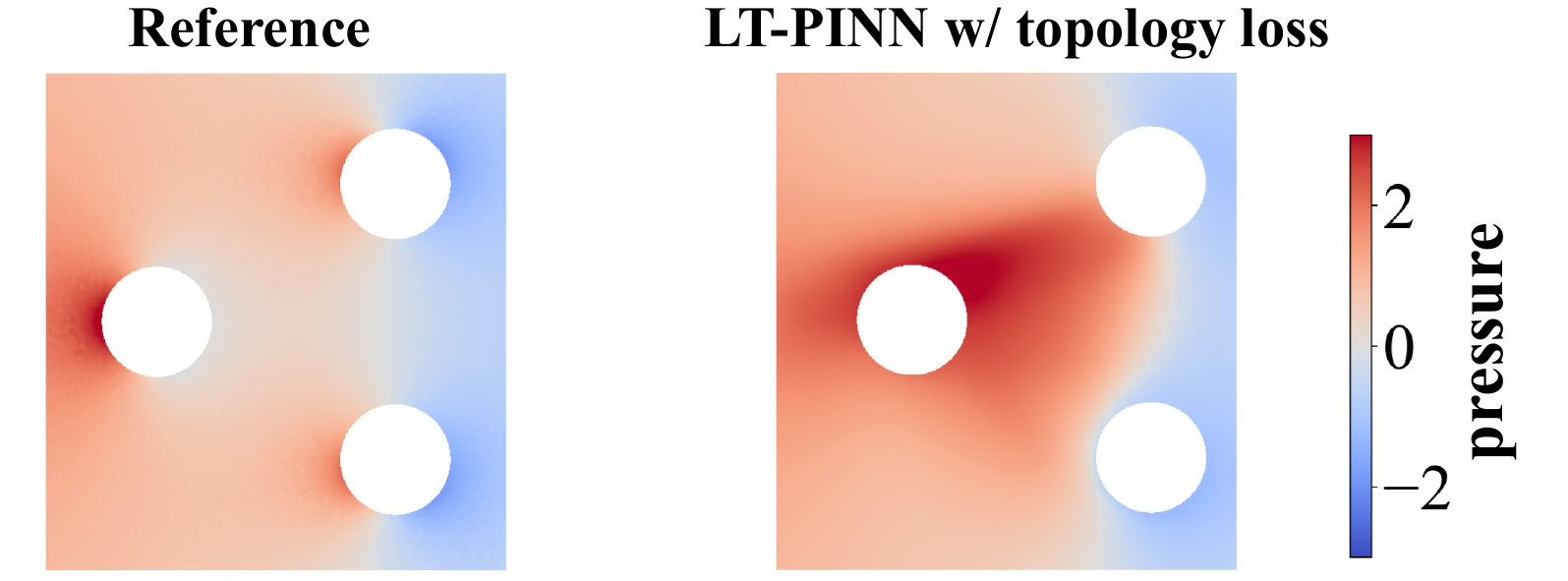}}
\subfigure[]{\label{fig:1d3c_field-b}
\includegraphics[width=0.7\textwidth]{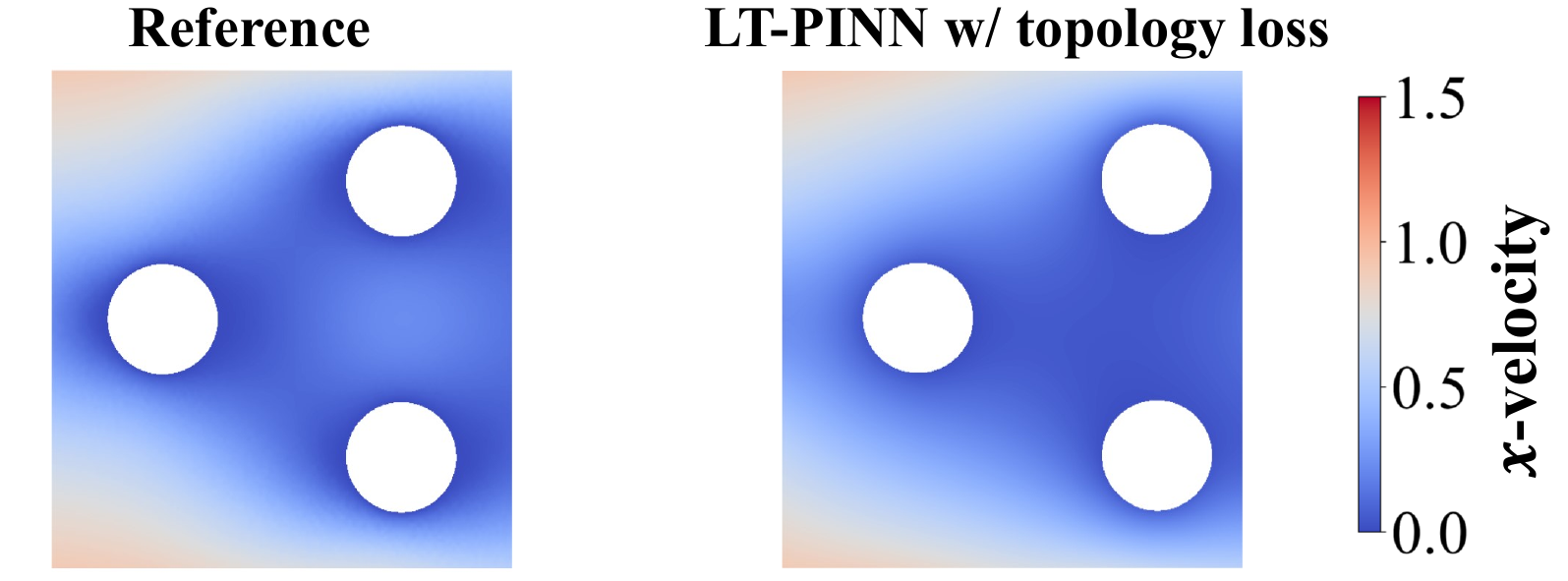}}
\subfigure[]{\label{fig:1d3c_field-c}
\includegraphics[width=0.7\textwidth]{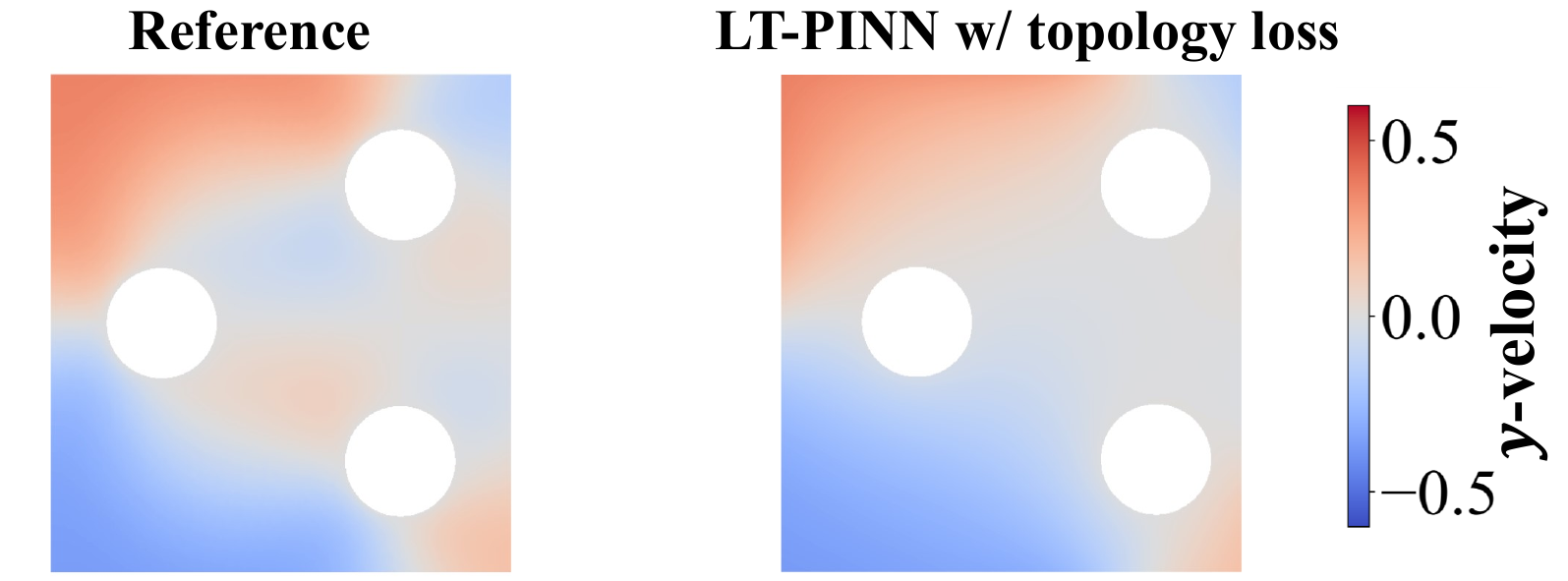}}
\caption{Time-independent flow: predicted velocity and pressure field via LT-PINN for 3-circle array: a) pressure; b) $x$-velocity; c) $y$-velocity.}\label{fig:1d3c_field}
\end{figure}

\begin{figure}[tp]
\centering%% For centre alignment of image.
\subfigure[]{\label{fig:1d8c_field-a}
\includegraphics[width=0.7\textwidth]{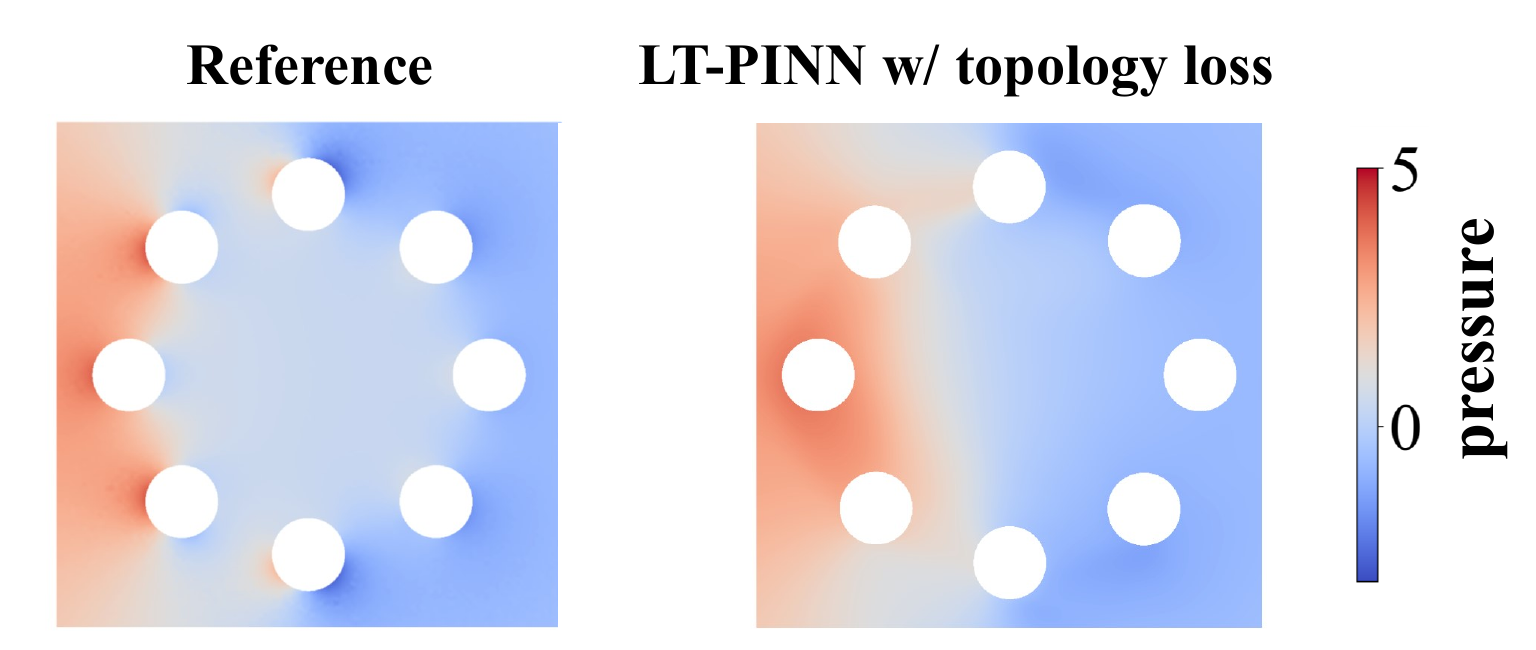}}
\subfigure[]{\label{fig:1d8c_field-b}
\includegraphics[width=0.7\textwidth]{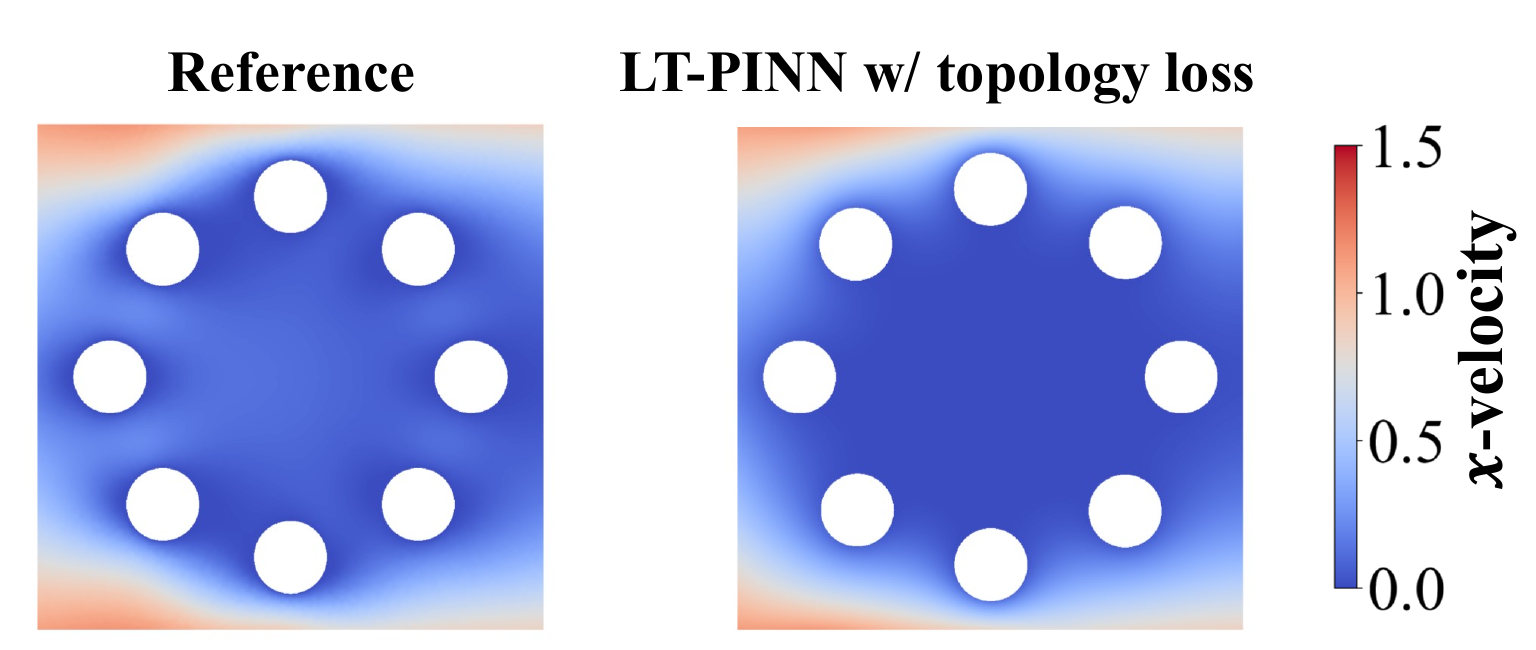}}
\subfigure[]{\label{fig:1d8c_field-c}
\includegraphics[width=0.7\textwidth]{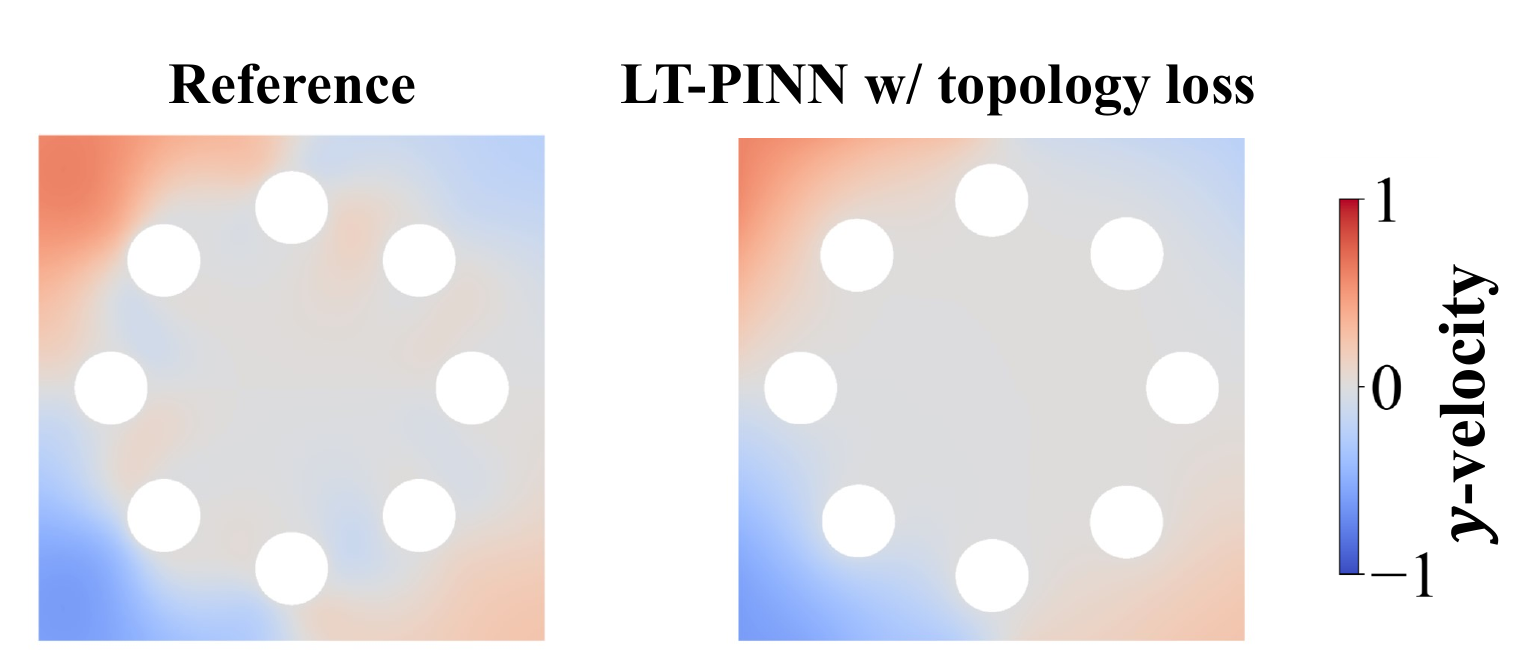}}
\caption{Time-independent flow: predicted velocity and pressure field via LT-PINN for 8-circle array: a) pressure; b) $x$-velocity; c) $y$-velocity.}\label{fig:1d8c_field}
\end{figure}

The final predicted velocity and pressure fields with the optimized topology via LT-PINNs are presented in Figs.\ref{fig:1d2c_field}, \ref{fig:1d3c_field}, and \ref{fig:1d8c_field}. The results show excellent agreement between predicted and reference fields. For the 8-circle array configuration (Fig. \ref{fig:1d8c_field-b}), the velocity field prediction successfully resolves the low-velocity zone in the central region, in close accordance with the reference solution. The pressure field prediction (Fig. \ref{fig:1d8c_field-a}) similarly exhibits high fidelity, accurately reproducing the characteristic pressure gradient from higher left-side values to lower right-side pressures, precisely matching the reference distribution. 

\begin{figure}[ht]
\centering%% For centre alignment of image.
\subfigure[]{\label{fig:steady_4-a}
\includegraphics[width=0.36\textwidth]{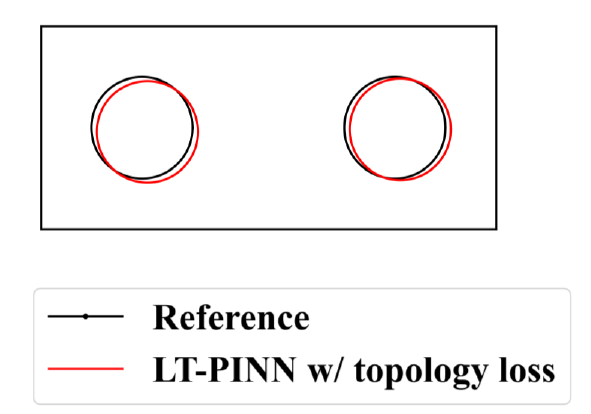}}
\hfil
\subfigure[]{\label{fig:steady_4-b}
\includegraphics[width=0.28\textwidth]{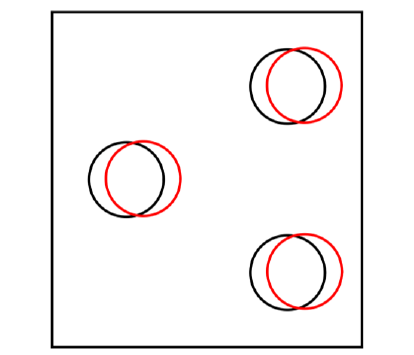}}
\hfil
\subfigure[]{\label{fig:steady_4-c}
\includegraphics[width=0.3\textwidth]{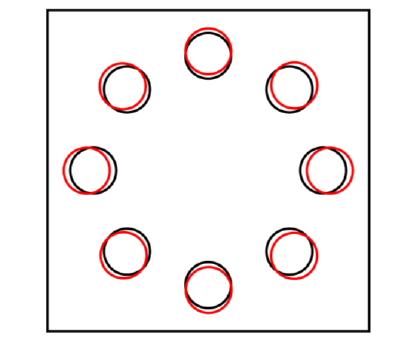}}
\caption{Time-independent flow: predicted topology via LT-PINNs w/ topology loss: a) 2-circle array; b) 3-circle array; c) 8-circle array.}\label{fig:steady_4}
\end{figure}

Across all test cases, LT-PINNs effectively reproduce the key fluid dynamic phenomena of flow diversion around multi-circle arrays. However, minor discrepancies persist between predicted and reference solutions. Notably, in the 3-circle array configuration (Fig. \ref{fig:1d3c_field-a}), the predicted pressure field of the LT-PINN shows a high pressure anomaly in the central region. We attribute this deviation primarily to insufficient measurement data available during the training phase, which may have limited the model's ability to fully resolve the complex flow interactions. Although minor inaccuracies persist in the physical field predictions, the optimized topological configurations demonstrate excellent agreement with reference solutions in Fig. \ref{fig:steady_4}. This finding indicates that the LT-PINN ensures robust performance in complex topology optimization tasks.

Furthermore, the error of the predicted velocity and pressure fields is estimated by the relative $L_2$ error of $128 \times 128$ uniformly distributed test sample points in the region (excluding those inside circles), i.e., $[L_{x,min}-D,L_{x,max}+D]\times[L_{y,min}-D,L_{y,max}+D]$, as listed in Tab. \ref{tab:L2-steady}.

As expected, the relative $L_2$ errors for multi-circle arrays are higher than those observed for the single-circle case (Sec. \ref{sec:compare}), where the governing PDEs are significantly simpler. However, since the relative $L_2$ error can be disproportionately amplified when the field's absolute mean value is small, we additionally employ the normalized mean absolute error (NMAE) \cite{lin2025towards} as a more robust error metric. The NMAE mitigates bias induced by low-magnitude reference fields and provides a clearer assessment of prediction accuracy.

The NMAE results, presented in Table \ref{tab:NMAE-steady}, illustrate that the maximum error occurs in the pressure prediction for the 3-circle array (NMAE = 0.1025), corroborating the earlier observation of elevated pressure deviations in Fig. \ref{fig:1d3c_field-a}. Nevertheless, the remaining NMAE values remain below 0.1, indicating that the predicted velocity and pressure fields generally deviate by less than 10\% relative to the reference solution. For practical engineering applications, this level of agreement illustrates sufficient accuracy, confirming the reliability of LT-PINNs for time-independent flow field prediction in complex topological configurations.

\begin{table}[ht!]
\centering
\begin{tabular}{c c c c}
\hline
\textbf{Relative $L_2$ error}  & \textbf{2-circle}& \textbf{3-circle}& \textbf{8-circle}\\ \hline
Pressure & 0.3995 & 0.9656 & 0.4253 \\ 
$x$-velocity& 0.2282 & 0.1620 & 0.1919 \\ 
$y$-velocity& 0.3151 & 0.3421 & 0.2849 \\ 
\hline
\end{tabular}
\caption{Time-independent flow: relative $L_2$ error of the predicted velocity and pressure via LT-PINN w/ topology loss.}
\label{tab:L2-steady}
% \end{sidewaystable}
\end{table}

\begin{table}[ht]
\centering
\begin{tabular}{c c c c}
\hline
\textbf{NMAE}  & \textbf{2-circle}& \textbf{3-circle}& \textbf{8-circle}\\ \hline
Pressure & 0.0387 & 0.1025 & 0.0559 \\ 
$x$-velocity& 0.0722 & 0.0533 & 0.0510 \\ 
$y$-velocity& 0.0391 & 0.0634 & 0.0335 \\ 
\hline
\end{tabular}
\caption{Time-independent flow: normalized mean absolute error (NMAE) of predicted velocity and pressure via LT-PINN w/ topology loss.}
\label{tab:NMAE-steady}
% \end{sidewaystable}
\end{table}

\subsubsection{Time-Dependent Flow Problem}
\label{sec:unsteady}
We then consider a even more non-linear case, time-dependent Navier-Stokes for flow problem, characterized as:
\begin{equation}\label{eq:unsteadyNS}
\begin{aligned}
\nabla \cdot \bm{u} &~= 0, \\
\bm{u}_t+(\bm{u} \cdot \nabla)\bm{u} &~= -\nabla{p} + \frac{1}{Re}\Delta{\bm{u}},\\
\bm{u} \bigg|_{bc} &= 0,
\end{aligned}
\end{equation}
where $\bm{u}_t$ is the gradient of velocity to time.

Since the circles are stationary, there is no need to infer the circle boundary change over time. Therefore, we omit the time terms in the Navier-Stokes equation by converting it into the pressure Poisson equation: 
\begin{equation}\label{eq:Poisson}
\begin{aligned}
\Delta p &~= -\nabla \cdot (\bm{u} \otimes \bm{u}), \\
\bm{u} \bigg|_{bc} &= 0,
\end{aligned}
\end{equation}
where $\otimes$ is the outer product.

For time-dependent flow problems, we employ $Re=100$ in Eq. \eqref{eq:unsteadyNS} to capture unsteady vortex shedding behind circular structures \cite{pingjian2009numerical}, while maintaining the same ROI edge conditions as the steady-state case described in Sec. \ref{sec:steady} for data generation. The LT-PINN formulation employs the instantaneous pressure Poisson equation (Eq. \eqref{eq:Poisson}) in its PDE loss function (Eq. \eqref{losseq2}), eliminating the need for time-resolved training data. Spatial training data at a time instance are sampled using the same methodology as the time-independent case (Sec. \ref{sec:steady}), with identical initialization procedures for the learnable parameters $\gamma$. To maintain focus, we evaluate only the most challenging 8-circle array configuration for time-dependent analysis. Complete details of data preparation and PINN training configurations are provided in Tab. \ref{tab:apend}. 

\begin{figure}[tp]
\centering
\begin{tabular}{cc}
\begin{minipage}[b]{0.38\textwidth}
    \centering
    \subfigure{\label{fig:1d8c_Re100-a}
    \includegraphics[width=\textwidth]{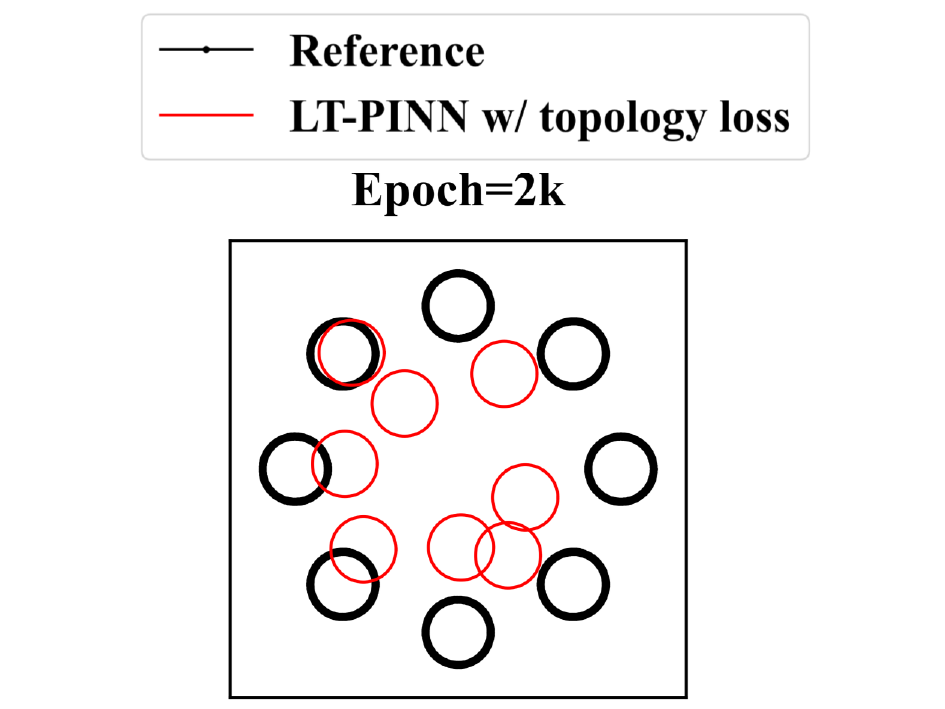}}
    (a)
\end{minipage}
&
\begin{minipage}[b]{0.35\textwidth}
    \centering
    \subfigure{\label{fig:1d8c_Re100-b}
    \includegraphics[width=\textwidth]{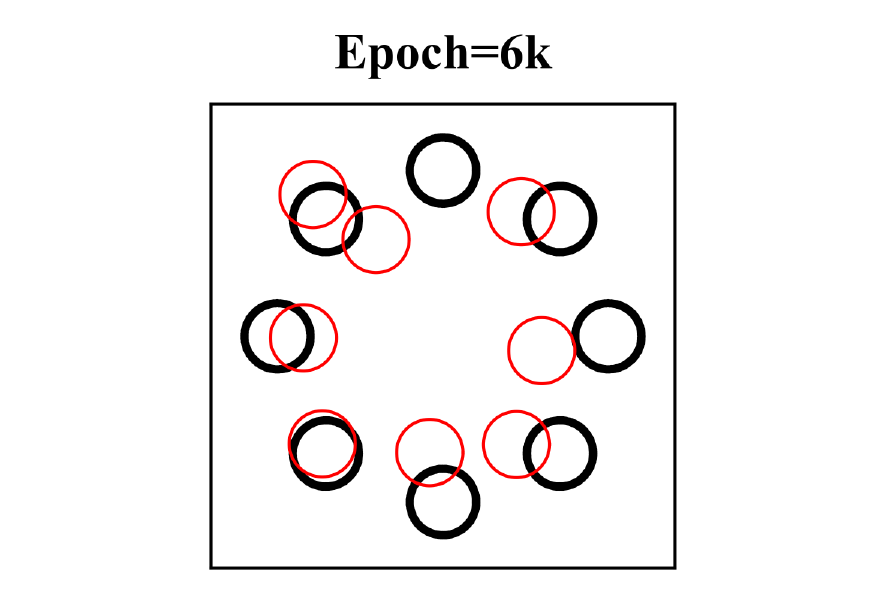}}
    (b)
\end{minipage}
\\
\multirow{2}{*}{
\begin{minipage}[t]{0.456\textwidth}
    \centering
    \subfigure{\label{fig:1d8c_Re100-e}
    \includegraphics[width=\textwidth]{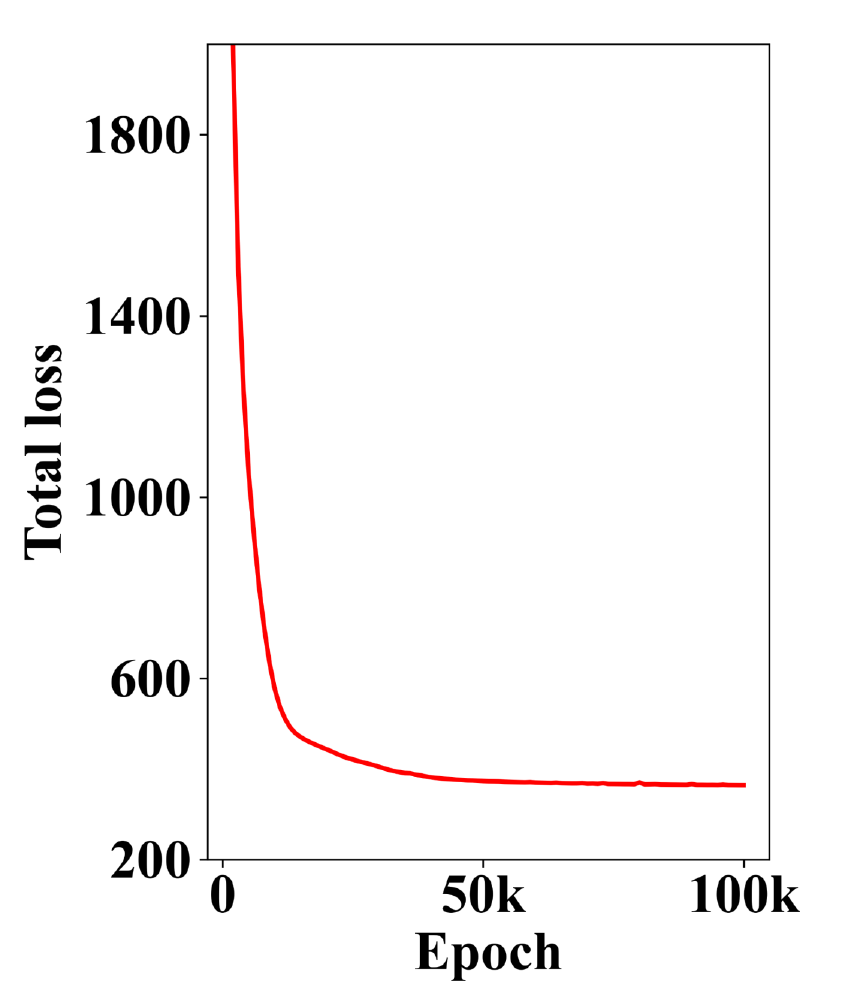}}
    (e)
\end{minipage}}
&
\begin{minipage}[t]{0.35\textwidth}
    \centering
    \subfigure{\label{fig:1d8c_Re100-c}
    \includegraphics[width=\textwidth]{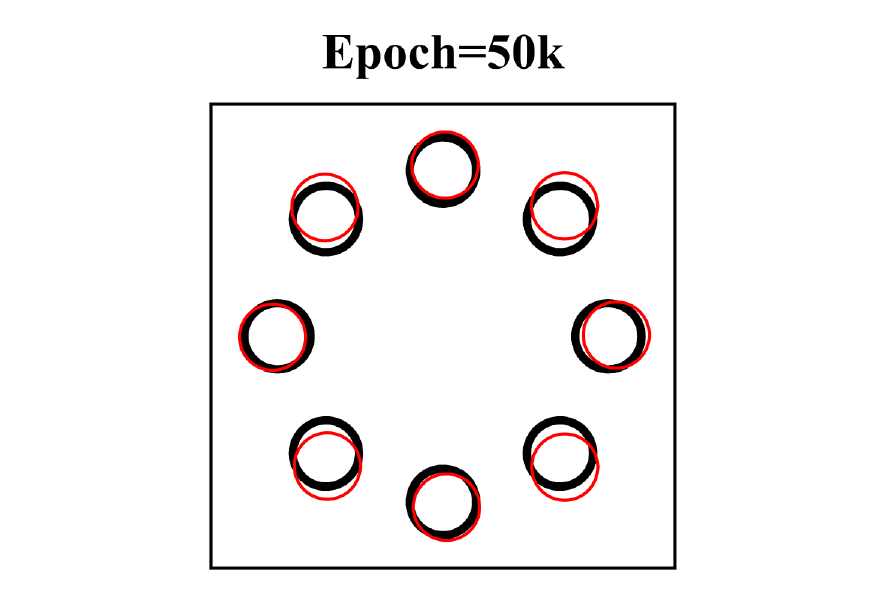}}
    (c)
\end{minipage} 
\\
&
\begin{minipage}[t]{0.35\textwidth}
    \centering
    \subfigure{\label{fig:1d8c_Re100-d}
    \includegraphics[width=\textwidth]{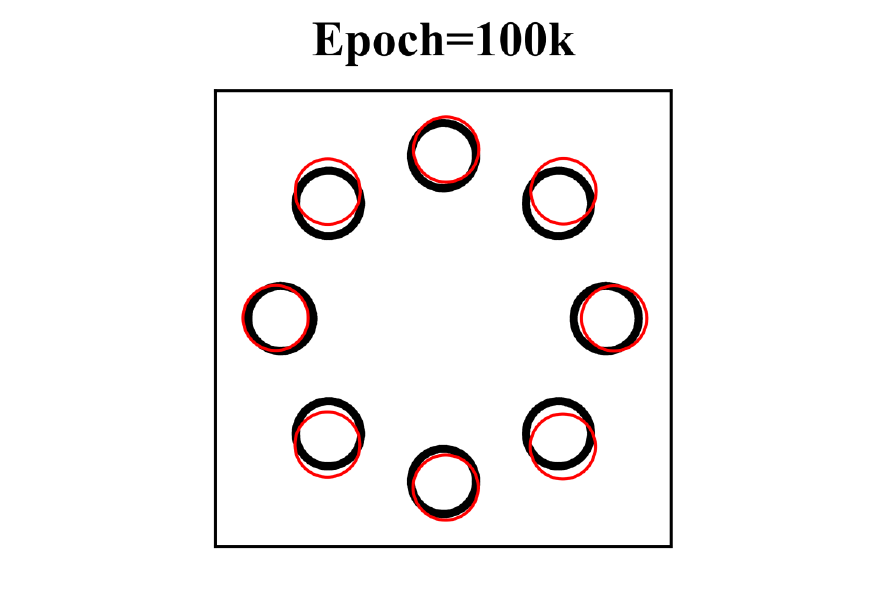}}
    (d)
\end{minipage} 
\end{tabular}
\caption{Time-dependent flow: loss history and $\gamma$ trajectory for 8-circle array: a-d) $\gamma$ trajectory at different training epochs; e) loss history.}
\label{fig:1d8c_Re100}
\end{figure}

To better understand the training process, the training loss history and $\gamma$ trajectory of 8-circle array is shown in Fig. \ref{fig:1d8c_Re100}. Upon reaching approximately 50k epochs, both the loss and predicted topology achieve stability. It is worth mentioning that the final predicted topology exhibits strong agreement with the reference configuration, validating the desired accuracy of the LT-PINN to infer topology in time-dependent flow problem.

\begin{figure}[tp]
\centering%% For centre alignment of image.
\subfigure[]{\label{fig:1d8c_field_Re100-a}
\includegraphics[width=0.9\textwidth]{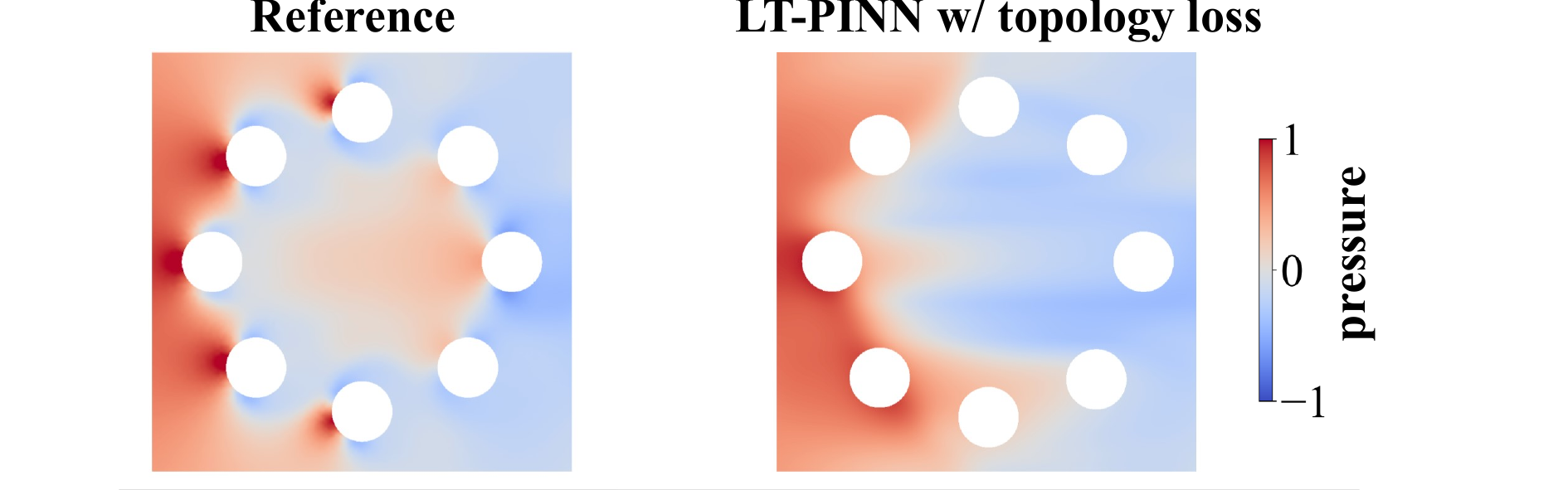}}
\subfigure[]{\label{fig:1d8c_field_Re100-b}
\includegraphics[width=0.9\textwidth]{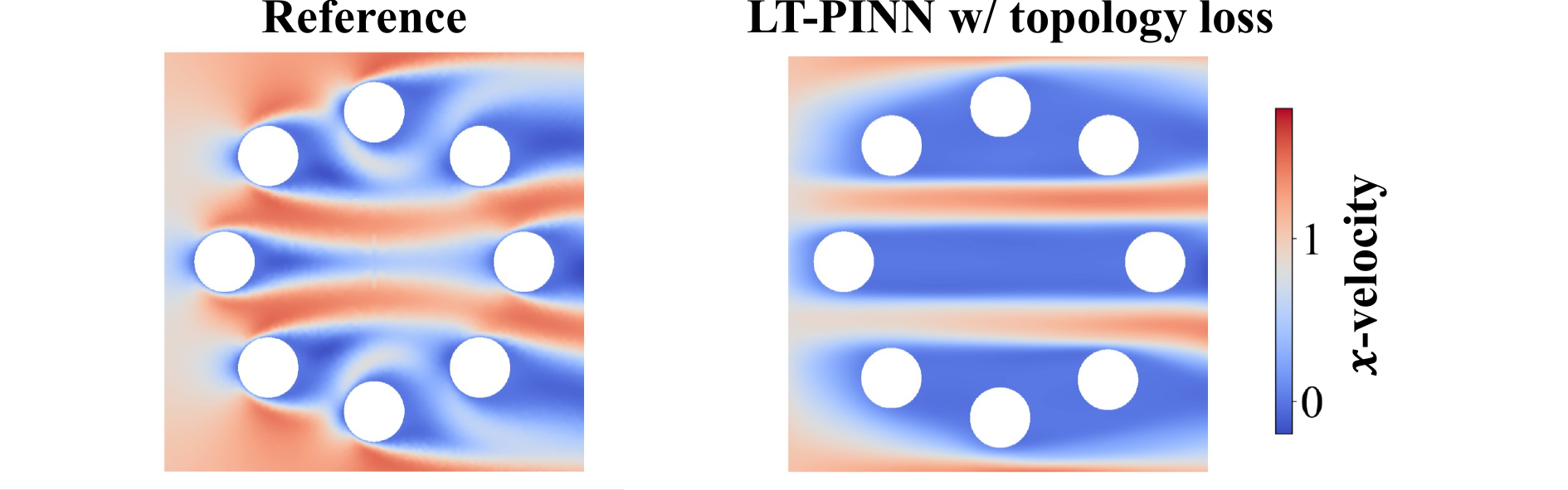}}
\subfigure[]{\label{fig:1d8c_field_Re100-c}
\includegraphics[width=0.9\textwidth]{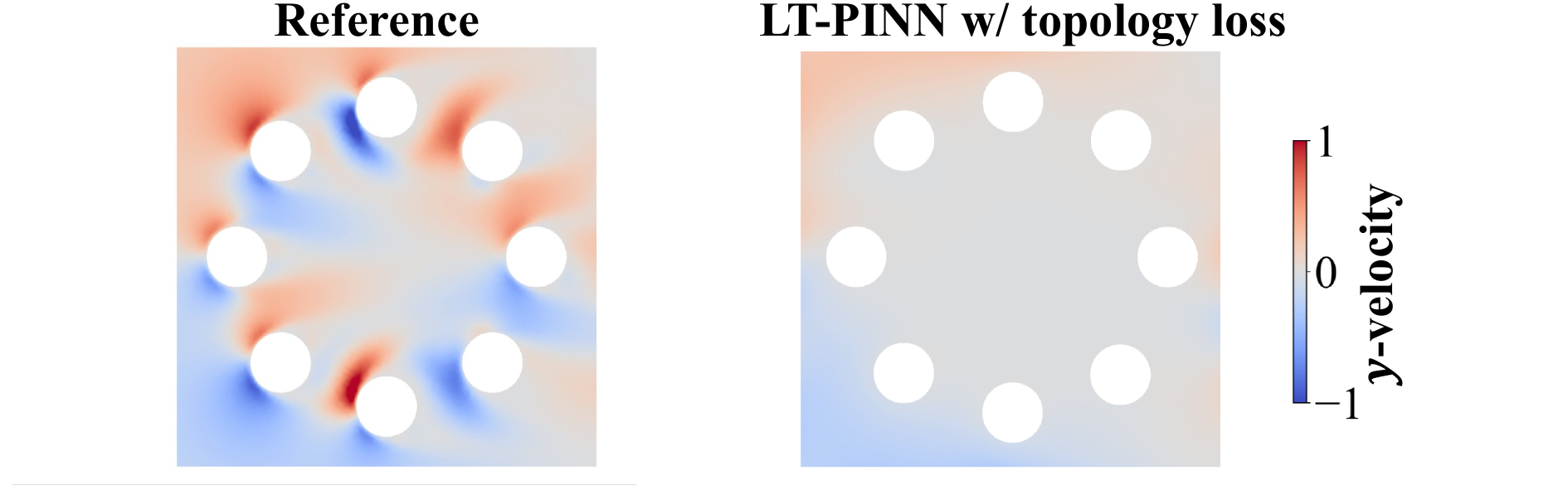}}
\caption{Time-dependent flow: predicted velocity and pressure field via LT-PINN for 8-circle array: a) pressure; b) $x$-velocity; c) $y$-velocity.}\label{fig:1d8c_field_Re100}
\end{figure}

The predicted velocity and pressure fields for the time-dependent flow case are presented in Fig. \ref{fig:1d8c_field_Re100}. Comparative analysis reveals generally good agreement between the LT-PINN predictions and the reference solution. Specifically, the pressure field prediction successfully reproduces the characteristic high-pressure region upstream of the multi-circle array and the corresponding low-pressure wake region, demonstrating consistency with the reference data. Regarding velocity field predictions, the LT-PINN accurately captures the formation of two high-velocity jets through the array, as evidenced in Fig. \ref{fig:1d8c_field_Re100-b}. However, certain flow features exhibit discrepancies. Most notably, while the reference solution shows curved jet trajectories with significant vertical velocity components, the LT-PINN prediction yields straighter jet paths with more uniform vertical velocity distributions (Fig. \ref{fig:1d8c_field_Re100-c}). The observed differences in flow field predictions, particularly in the vertical velocity components, indicate potential areas for model improvement. We hypothesize that the inclusion of additional measurement data within the domain would enhance the accuracy of flow pattern predictions.   

The hydrodynamic forces on the circles provide critical quantitative measures of flow behavior. We assess predictive performance by computing pressure-induced lift and drag forces for the 8-circle array (Table \ref{tab:Lift}). The results show good agreement in lift force predictions, but a substantial 62.97\% underestimation of drag forces. As a result, this discrepancy produces a corresponding 1.677-fold overestimation of the lift-to-drag ratio relative to the reference solution. 

\begin{table}[ht]
\centering
\begin{tabular}{c c c}
\hline
\textbf{Force on multi-circle array}  &  \textbf{Reference} &  \textbf{LT-PINN}\\ \hline
Lift & 0.0578 & 0.0574\\ 
Drag & 3.7337 & 1.3827  \\ 
Lift-drag ratio & 0.0155 & 0.0415 \\ 
\hline
\end{tabular}
\caption{Time-dependent flow: predicted lift, drag, and lift-drag ratio via LT-PINN.}
\label{tab:Lift}
% \end{sidewaystable}
\end{table}

Given the observed discrepancies in both flow patterns and force predictions for the 8-circle array configuration, we employ the NMAE as a metric to systematically evaluate the predictive accuracy of the LT-PINN. The NMAE is calculated on the test sample points extracted from the ROI as described in Sec. \ref{sec:steady}, and results are listed in Tab. \ref{tab:NMAE-unsteady}. The analysis indicates that while the $x$-velocity component exhibits a relatively elevated NMAE of 0.2032 due to unresolved fine-scale flow features between adjacent circles, both pressure and $y$-velocity predictions achieve satisfactory accuracy (errors \textless 10.2\%) within standard engineering tolerances. These results demonstrate LT-PINN's capability for concurrent topology optimization and PDE solution prediction in time-dependent flows. 
\begin{table}[ht]
\centering
\begin{tabular}{c c}
\hline
\textbf{NMAE}  &  \textbf{Time-dependent flow}\\ \hline
Pressure & 0.1011\\ 
$x$-velocity& 0.2032  \\ 
$y$-velocity& 0.0595 \\ 
\hline
\end{tabular}
\caption{Time-dependent flow: NMAE of the predicted velocity and pressure via LT-PINN w/ topology loss.}
\label{tab:NMAE-unsteady}
% \end{sidewaystable}
\end{table}

\subsection{Application on Flow Velocity Rearrangement}
\label{sec:rearrange}
The upstream flow velocity profile represents a critical characteristic influencing numerous flow phenomena, including turbulence development \cite{george2012asymptotic}, flow measurement accuracy \cite{morrison1994upstream}, biofilm growth in water distribution systems \cite{tsai2005impact}, and gas cyclone separator performance \cite{zhou2018massive}. Consequently, controlled manipulation of upstream velocity profiles to achieve target downstream velocity profiles demonstrates substantial engineering potential.  

In this section, we formulate a flow velocity rearrangement task for illustrating the practical feasibility of the LT-PINN, where the LT-PINN can generate an optimal topology based solely on the upstream velocity profile without prior topological knowledge, to achieve a specified target downstream velocity profile. In detail, the ROI is expanded to be $25D\times15D$. Boundary conditions consist of a uniform inlet velocity profile at the left edge ($x$-velocity $=1$, $y$-velocity $=0$), periodic conditions at the top and bottom edges, and a target downstream $x$-velocity profile $u=sin(2\pi x/15)+1$ at the right edge, with Reynolds number $Re=1$. Data loss function are implemented to enforce both the inlet/outlet velocity profiles and the periodicity condition between the top and bottom edges, while no pressure-based data loss is applied. For topology optimization, LT-PINN utilizes 48 circles with allowed overlap, providing sufficient design freedom without topology loss constraints in this configuration. Since the ROI in this case is much larger than that of other cases, resulting in a larger amount of sampled data, we employ 4xGPUs as parallel computing resource for this case. Complete details regarding data sampling and PINN training parameters are provided in Tab. \ref{tab:apend}. 

\begin{figure}[ht]
\centering
\begin{tabular}{cc}
\begin{minipage}[b]{0.45\textwidth}
    \centering
    \subfigure{\label{fig:conversion_topology-a}
    \includegraphics[width=\textwidth]{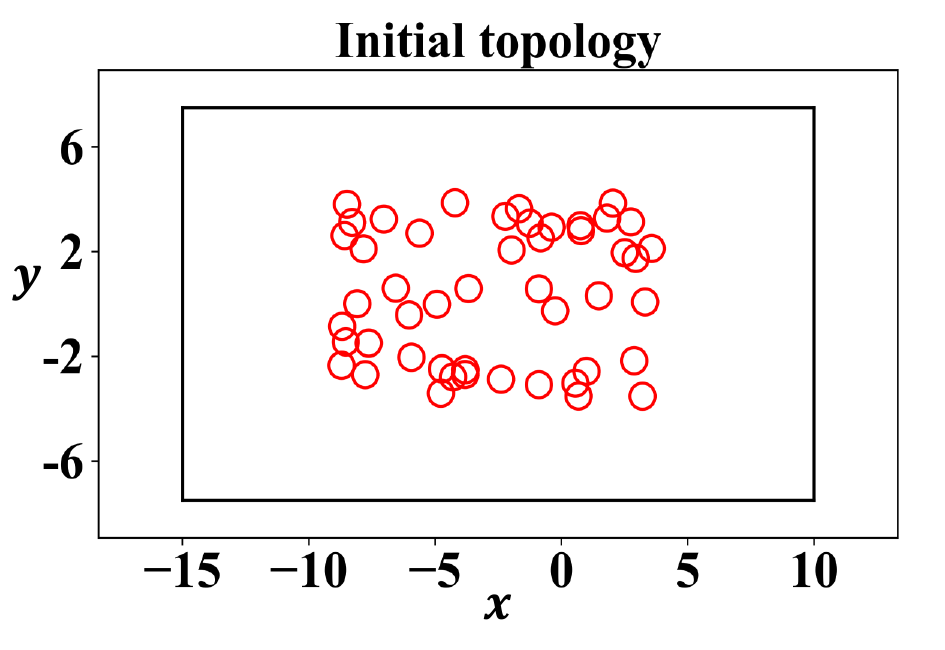}}
    (a)
\end{minipage}
&
\begin{minipage}[b]{0.45\textwidth}
    \centering
    \subfigure{\label{fig:conversion_topology-b}
    \includegraphics[width=\textwidth]{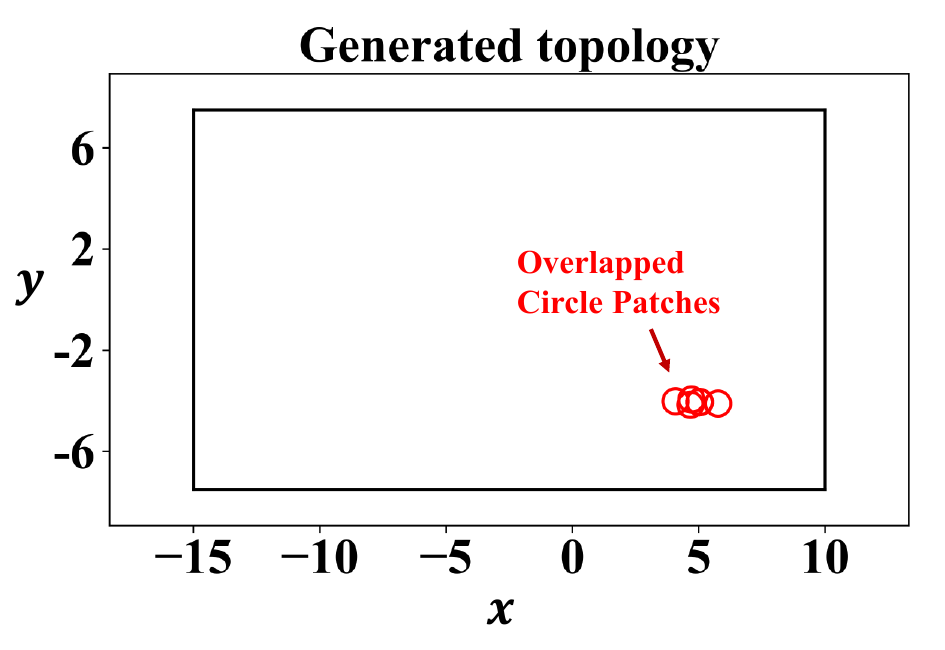}}
    (b)
\end{minipage}
\\
\begin{minipage}[t]{0.45\textwidth}
    \centering
    \subfigure{\label{fig:conversion_topology-c}
    \includegraphics[width=\textwidth]{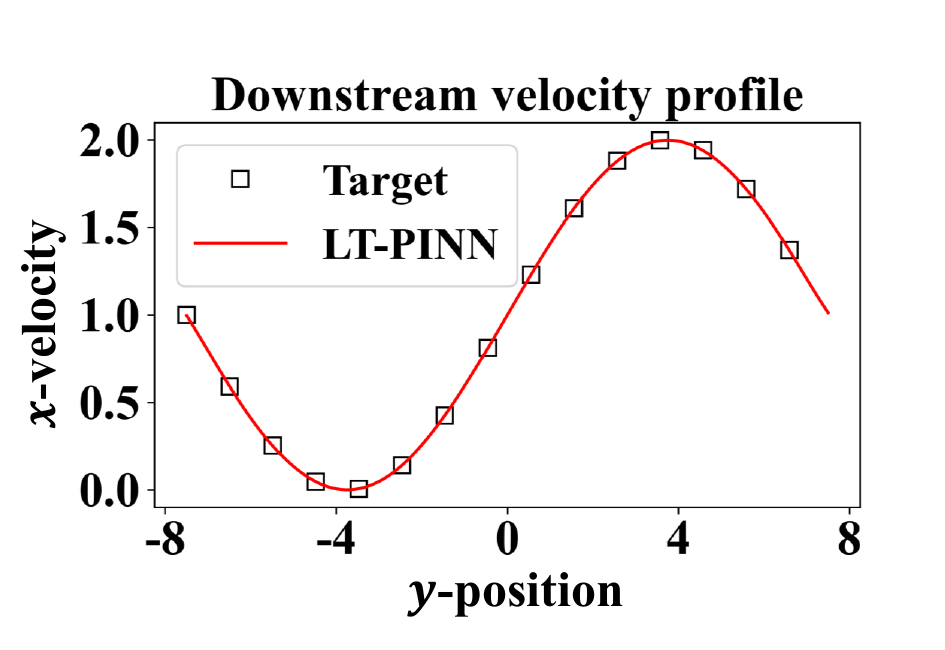}}
    (c)
\end{minipage}
\end{tabular}
\caption{Flow rearrangement: generated topology and rearranged downstream velocity profile via LT-PINN: a) initial topology; b) generated topology; c) rearranged downstream velocity profile.}
\label{fig:conversion_topology}
\end{figure}

Fig. \ref{fig:conversion_topology} shows the optimized topology and resulting downstream velocity profile from LT-PINN prediction. Initially, 48 randomly distributed circles populate the domain (Fig. \ref{fig:conversion_topology-a}). During optimization, these patches undergo substantial spatial reorganization, ultimately coalescing into an intricate cluster in the bottom-right region (Fig. \ref{fig:conversion_topology-b}). The cluster's exact configuration can be precisely determined, since each circle's position is explicitly defined. Furthermore, as demonstrated in Fig. \ref{fig:conversion_topology-c}, the downstream $x$-velocity profile successfully achieves the target sinusoidal distribution.

\begin{figure}[ht]
\centering%% For centre alignment of image.
\includegraphics[width=0.8\textwidth]{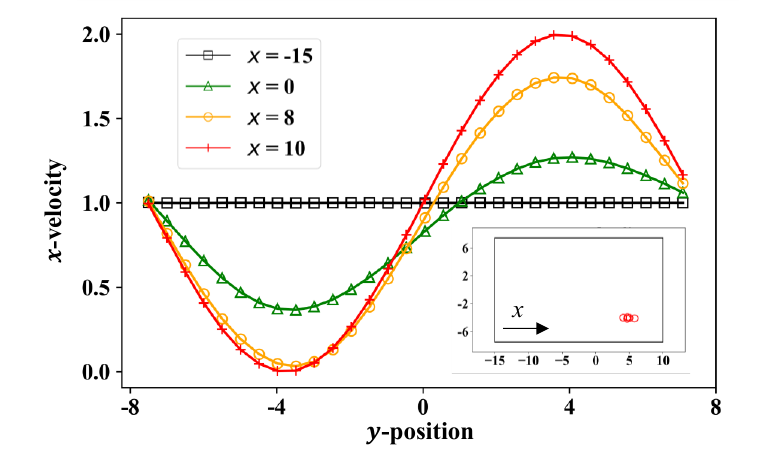}
\caption{Flow rearrangement: predicted velocity at different $x$-positions via LT-PINN.}\label{fig:conversion_velocityprofile}
\end{figure}

The conversion of the detailed $x$-velocity profile from a uniform upstream velocity to a target sine-shaped downstream velocity is depicted in Fig. \ref{fig:conversion_velocityprofile}. It is evident that the $x$-velocity profile conversion occurs more rapidly in the lower $y$-region, completing by $x=8$, while the upper region requires a longer development length. This asymmetric conversion pattern results from the topology's cluster configuration in the bottom-right corner (centered near $x=5$). 

\begin{figure}[tp]
\centering%% For centre alignment of image.
\subfigure[]{\label{fig:conversion_field-a}
\includegraphics[width=1\textwidth]{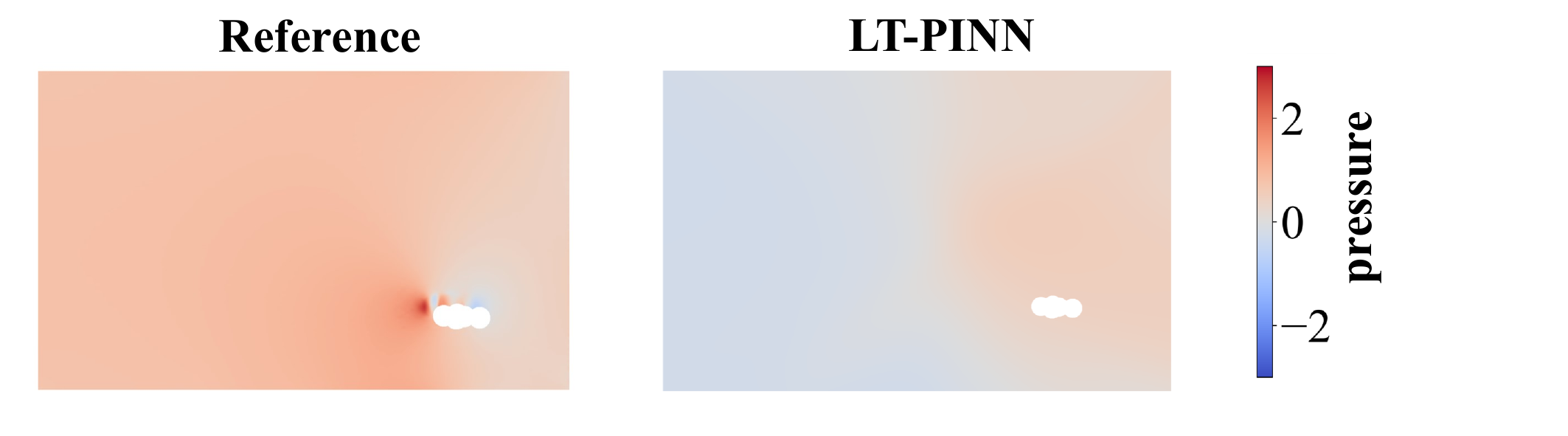}}
\subfigure[]{\label{fig:conversion_field-b}
\includegraphics[width=1\textwidth]{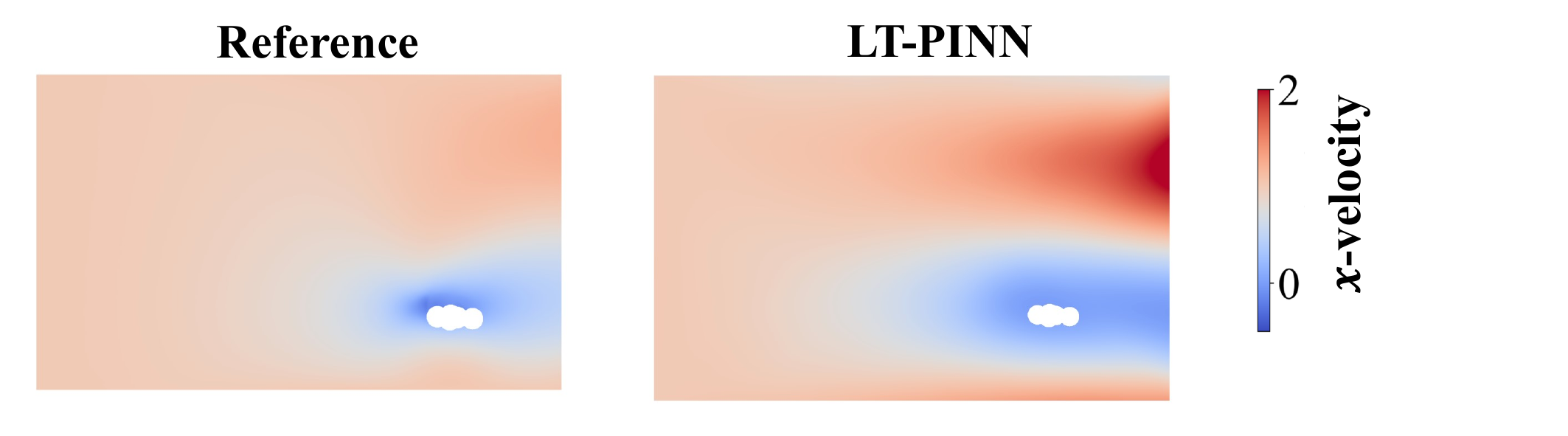}}
\subfigure[]{\label{fig:conversion_field-c}
\includegraphics[width=1\textwidth]{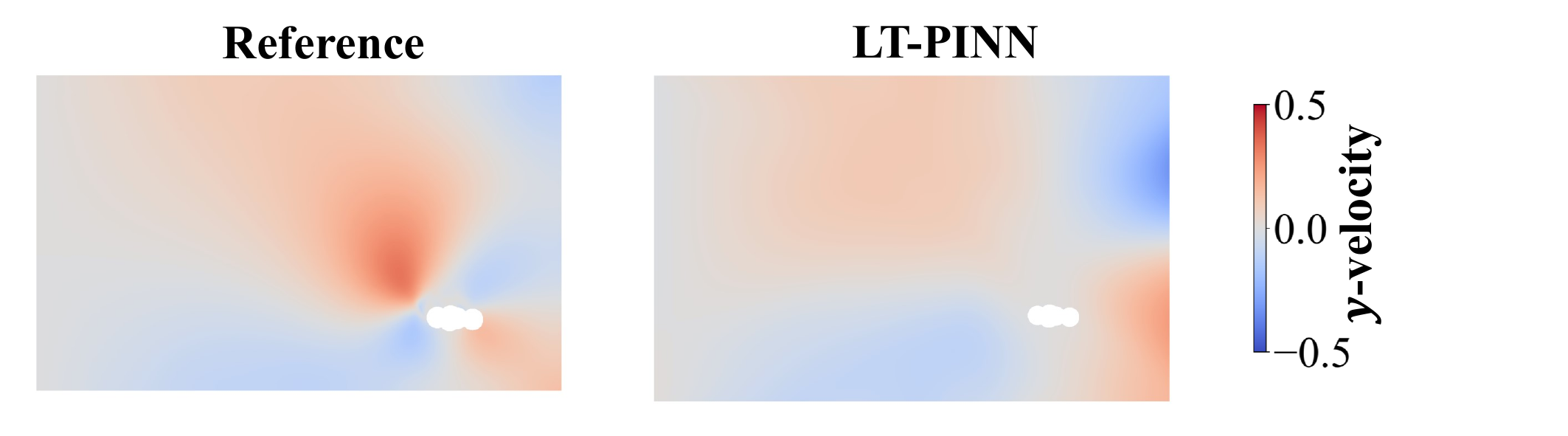}}
\caption{Flow rearrangement: predicted velocity and pressure field via LT-PINN: a) pressure; b) $x$-velocity; c) $y$-velocity.}\label{fig:conversion_field}
\end{figure}

Based on the cluster topology generated by the LT-PINN, a Computer-Aided Design (CAD) model is constructed to facilitate Computational Fluid Dynamics (CFD) simulations for the purpose of demonstrating the compatibility of the LT-PINN with CAD and CFD, in addition to generate reference validation data. Equivalent velocity boundary conditions are applied to the left, top, and bottom edges with a corresponding pressure boundary condition on the right edge. Under the prescribed boundary conditions, a comprehensive comparison between the predicted flow fields and reference solutions is presented in Fig. \ref{fig:conversion_field}. As observed, LT-PINN successfully captures the low $x$-velocity region downstream of the cluster, demonstrating good agreement with the reference data. However, notable discrepancies are observed in several aspects: (1) The $x$-velocity profile near the right edge exhibiting a more pronounced gradient between the high-velocity (top-right) and low-velocity (bottom-right) regions in LT-PINN predictions compared to the reference solution (Fig. \ref{fig:conversion_field-b}); and (2) Evident differences in both pressure and $y$-velocity fields, particularly upstream of the cluster where the reference solution shows elevated pressure and $y$-velocity magnitudes relative to the LT-PINN predictions. 

\begin{figure}[tp]
\centering
\begin{tabular}{cc}
\begin{minipage}[b]{0.45\textwidth}
    \centering
    \subfigure{\label{fig:conversion_bcs-a}
    \includegraphics[width=\textwidth]{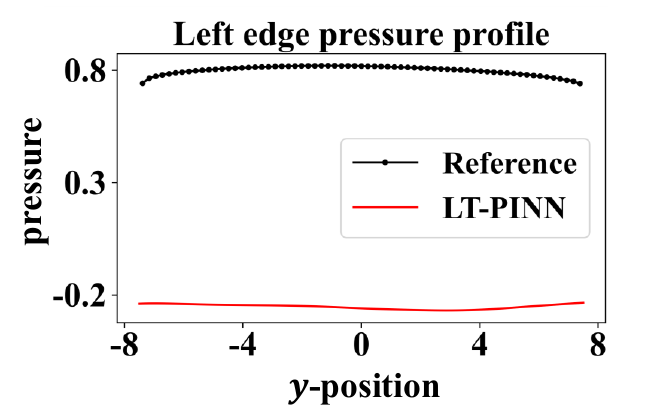}}
    (a)
\end{minipage}
&
\begin{minipage}[b]{0.45\textwidth}
    \centering
    \subfigure{\label{fig:conversion_bcs-b}
    \includegraphics[width=\textwidth]{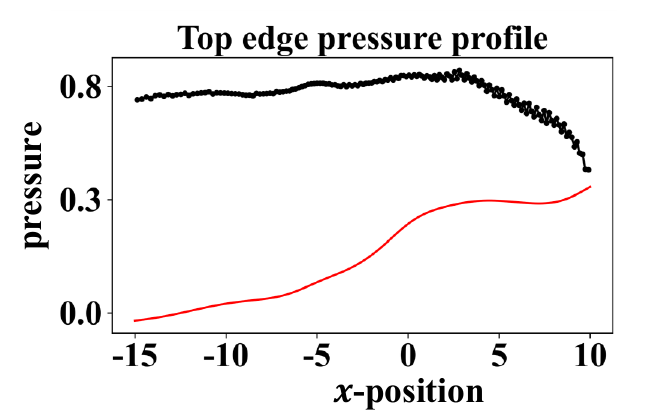}}
    (b)
\end{minipage}
\\
\begin{minipage}[t]{0.45\textwidth}
    \centering
    \subfigure{\label{fig:conversion_bcs-c}
    \includegraphics[width=\textwidth]{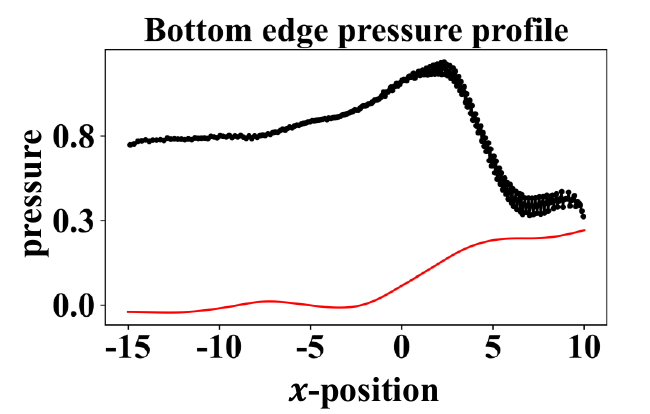}}
    (c)
\end{minipage}
&
\begin{minipage}[t]{0.45\textwidth}
    \centering
    \subfigure{\label{fig:conversion_bcs-d}
    \includegraphics[width=\textwidth]{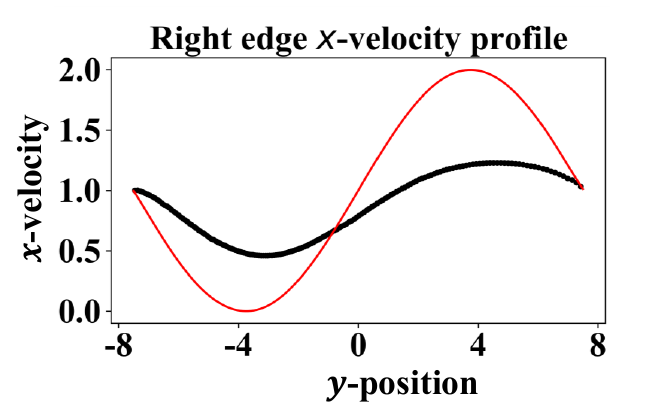}}
    (d)
\end{minipage}
\\
\begin{minipage}[t]{0.45\textwidth}
    \centering
    \subfigure{\label{fig:conversion_bcs-e}
    \includegraphics[width=\textwidth]{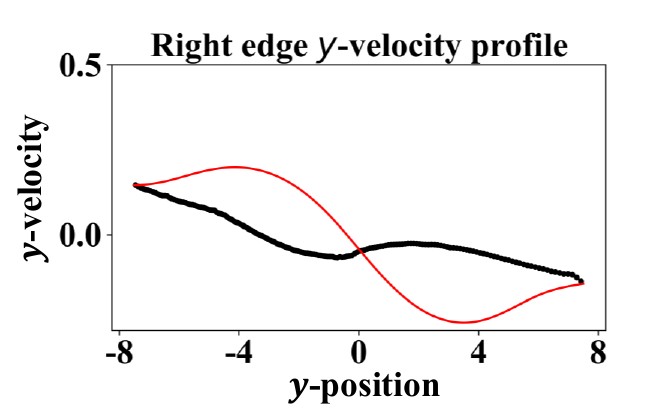}}
    (e)
\end{minipage}
\end{tabular}
\caption{Flow rearrangement: comparison between the LT-PINN and reference on the undefined edge conditions: a, b, c) are undefined pressure edge conditions; d, e) are undefined velocity edge conditions.}
\label{fig:conversion_bcs}
\end{figure}

The above discrepancies are hypothesized to originate from differences in imposing four edges' conditions between the LT-PINN and the reference solution. As illustrated in Fig. \ref{fig:conversion_bcs}, the undefined edge conditions, comprising velocity specifications at the right edge and pressure conditions along the remaining three edges, exhibit notable variations. Quantitative analysis reveals substantial differences in the undefined pressure edge conditions, while the velocity edge conditions at the right edge demonstrate relatively minor deviations.

Specifically, the reference solution's downstream $x$-velocity profile displays a sinusoidal pattern comparable to the LT-PINN results. Furthermore, the $y$-velocity components at $x=-7.5$ and $x=7.5$ show close agreement between both sets of results. These observations indicate that pressure edge condition differences constitute the primary source of flow field discrepancies between the reference solution and LT-PINN results. However, from the perspective of topological generation capability, particularly in converting uniform upstream velocity into the target sine-shape downstream velocity profile, LT-PINN predictions match with the reference solution, as illustrated in Fig. \ref{fig:conversion_bcs-d}.

\subsection{Key Insights}
\label{subsec:discussion}
This study presents a novel LT-PINN that simultaneously performs topology optimization and solves governing PDEs. By integrating the meshless advantages of PINNs with Lagrangian topology optimization techniques, LT-PINN offers an efficient, effective, and unified approach for boundary-focused engineering optimization problems. 

We first benchmark LT-PINNs against state-of-the-art DT-PINNs using two canonical test cases: (1) 2D elastic equations with Dirichlet boundary conditions and (2) 2D Laplace‘s equations with Neumann boundary conditions. The comparative analysis demonstrates LT-PINNs' superior accuracy, achieving a 66.53\% reduction in displacement prediction error (relative $L_2$ error) for elastic equations and a remarkable 99.42\% reduction relative $L_2$ error for predicted heat flux across the boundary for Laplace's equations compared to DT-PINNs. These results conclusively establish LT-PINNs' enhanced capability for PDE solution accuracy across different boundary condition types. Furthermore, LT-PINNs demonstrate superior topology inference accuracy compared to DT-PINNs, particularly for Laplace's equations with Neumann boundary conditions. While DT-PINNs produce irregular topological configurations due to their inability to properly incorporate Neumann conditions in their total loss function (Eq. \eqref{eq:dirichlet}), LT-PINNs achieve precise topology reconstruction through correct boundary condition encoding (Eq. \eqref{losseq2} ). Another key advantage of LT-PINNs is their automatic generation of well-defined topological boundaries, in contrast to DT-PINNs which require subjective manual interpolation of density thresholds. This automated approach eliminates a significant source of human-induced variability, ensuring more reliable and reproducible topology inference. 

Having validated LT-PINNs' effectiveness across different boundary conditions through comparison with DT-PINNs on fundamental test cases, we extend their performance investigation on more challenging applications: complex topology optimization problems and highly nonlinear Navier-Stokes equations for 2D flow around multi-circle arrays. Notably, LT-PINNs operate using only edge data from the four edges of ROI, without any interior measurements, presenting a significant computational challenge. For time-independent flow at $Re=1$, LT-PINNs accurately reconstruct 2-, 3-, and 8-circle array configurations, with velocity and pressure field predictions maintaining the NMAE below 11\% of each physical field's absolute range. In time-dependent flow ($Re=100$) past an 8-circle array, while topological inference remains accurate, velocity and pressure predictions show higher NMAE up to 20.32\%. However, LT-PINNs achieve satisfactory accuracy in predicting lift forces in the 8-circle array case. The future work may incorporate additional measurement data for LT-PINNs training to further improve prediction fidelity. 

Validation across 2- to 8-circle array configurations indeed confirms LT-PINNs' capability for complex topology optimization. However, those complex topology contains self-similar patterns, the characteristics of which are inherently encoded in the topology loss function. To further assess the generalizability of LT-PINNs in scenarios without such prior knowledge, we conduct an additional study focusing on a flow velocity rearrangement problem, a task of significant relevance to various downstream fluid dynamics applications. In this experiment, 48 randomly distributed circle patches are initialized within the computational domain for topology optimization. The objective is to evolve these patches into an optimal configuration capable of converting a uniform upstream velocity profile into a specified sinusoidal downstream velocity profile. Notably, the topology loss function is intentionally excluded from the optimization process, and the circle patches are allowed to overlap. As training progresses, the circle patches exhibit a tendency to coalesce, ultimately forming an irregular cluster topology near the downstream right edge of the domain. Despite the absence of explicit topological constraints, the resultant downstream velocity profile demonstrates strong agreement with the target sinusoidal distribution. While geometrically irregular, the explicit parameterization of each patch enables straightforward CAD reconstruction, demonstrating LT-PINNs' strong potential for engineering design integration. Complementary computational fluid dynamics (CFD) simulations are also performed to demonstrating the compatibility of LT-PINNs with CAD and CFD, and validate physical realizability of the inferred topology. The resulting downstream velocity profiles exhibits close alignment with the target sinusoidal distribution, further validating the efficacy of the optimized cluster topology in this flow velocity rearrangement task.

Throughout our numerical experiments presented in this study, we have limited our investigation to circular topological patches. Future work should explore extensions to arbitrary patch geometries with deformable boundaries described by higher-order parametric curves, for example B-spline curve. Besides, LT-PINN applications should be expanded to more realistic three-dimensional problems, such as metamaterial design \cite{bonfanti2024computational}, offshore structure optimization \cite{he2024obstacle}, sediment transport management \cite{you2025characterization}, and artificial reef optimization \cite{ronglan2024architected}. 

\section{Conclusion}
\label{sec:conclusion}
To address the critical need for simultaneous topology optimization and PDE solution, we present LT-PINNs, a novel and unified framework that combines topological design and optimization with PDE solving in a meshless Lagrangian formulation. Our approach demonstrates substantial improvements in both accuracy and versatility, surpassing current state-of-the-art counterpart, i.e., DT-PINNs. Its feasibility and engineering potential are further demonstrated through a series of experiments. The key innovations of this work include: 

\indent 1. LT-PINNs eliminate the error-prone manual boundary interpolation required in DT-PINNs, leading to a clearer and more efficient topology inference;

\indent 2. LT-PINNs precisely encode arbitrary boundary conditions through boundary condition loss function, resulting in a substantial reduction of the relative $L_2$ error in the predicted PDE solution;

\indent 3. For complex topology systems, LT-PINNs can incorporate topology loss function, particularly effective for self-similarity topology patterns;

\indent 4. For flow velocity rearrangement tasks, LT-PINNs can generate manufacturable topologies directly compatible with CAD, without the measurement data or prior knowledge of topological features.

Future work will focus on extending LT-PINNs to cope with more general topologies with deformable boundaries and 3-dimensional problems, thereby significantly broadening their applicability. 

\section*{Acknowledgments}
The research is financially supported by the start-up grant from The Hong Kong Polytechnic University and the research project funding from the Research Institute for Sustainable Urban Development (RISUD) at The Hong Kong Polytechnic University. In addition, HT would like to acknowledge financial support from Research Grants Council of Hong Kong under General Research Fund (15218421). Finally, we are grateful to Prof. Xin Bian and Mr. Yongzheng Zhu from Zhejiang Univeristy for their insightful discussions on PINNs.

\newpage
\appendix
\section{Computational Setup}\label{sec:details_computation}

In this section, the computational setup, including the architectures of neural networks, optimizer, etc., for all test cases, is presented for interested readers to replicate analysis. Details are summarized in the following table in addition to the case descriptions in Sec. \ref{sec:compare}.

\renewcommand{\thetable}{S\arabic{table}} % 重新定义表格序号格式
\setcounter{table}{0} % 重置表格计数器为0

\begin{sidewaystable}
%\begin{table}[ht]
\centering
\begin{tabular}{c c c c c c}
\hline
\textbf{Settings}  & \textbf{Sec. \ref{sec:steady}} & \textbf{Sec. \ref{sec:steady}} & \textbf{Sec. \ref{sec:steady}}   & \textbf{Sec. \ref{sec:unsteady}}  & \textbf{Sec. \ref{sec:rearrange}} \\  &\textbf{2-circle}& \textbf{3-circle}& \textbf{8-circle}& \textbf{8-circle} & \textbf{48-circle}\\
\hline
\# of layers & 5& 5& 5 & 5 & 5  \\ 
\# of neurons per layer & 64 & 64 & 64 & 64 & 64  \\ 
activation function & Tanh & Tanh& Tanh& Tanh & Tanh  \\ 
optimizer & Adam & Adam & Adam& Adam& Adam \\ 
learning rate & 0.0001 & 0.0001& 0.0001 & 0.0001& 0.0001  \\ 
training epoch & 100,000& 100,000& 100,000& 100,000& 100,000\\ 
$N$ & 270 $\times$ 120 & 249 $\times$ 270 & 420 $\times$ 420 & 420 $\times$ 420 &1500 $\times$ 900\\ 
$N_{b}$ & 1,024&1,536&4,096 & 4,096 & 20,480\\
$N_{d}$ & 3,389&3,682&5,373  & 5,373 & 800\\
$N_{t}$ & 2&3&8 & 8 & 48\\
$\lambda_{p}$ & $2 \times 10^3$ & $2 \times 10^3$& $2 \times 10^3$& $2\times10^3$ & $2\times10^3$  \\ 
$\lambda_{b}$ & $10^4$& $10^4$& $10^4$ & $10^4$ & $10^4$  \\ 
$\lambda_{d}$ & $10^4$ & $10^4$& $10^4$& $10^4$ & $10^4$  \\ 
$\lambda_{t}$ & $10^4$& $10^4$& $10^2$ & $10^2$ & -  \\ 
PDE &Eq. \eqref{eq:steadyNS} &Eq. \eqref{eq:steadyNS}&Eq. \eqref{eq:steadyNS}& Eq. \eqref{eq:Poisson} & Eq. \eqref{eq:steadyNS} \\
\# of topology patches & 2&3&8 & 8 & 48 \\
\# of GPUs & 1 & 1& 1& 1 & 4  \\ \hline
\end{tabular}
\caption{Details for the LT-PINNs in different test cases.}
\label{tab:apend}
\end{sidewaystable}
%\end{table}

\newpage
\bibliographystyle{unsrt}
\bibliography{ref} 
\end{document}